\documentclass{article} 
\usepackage{iclr2027_conference,times}

\usepackage{amsmath,amsfonts,bm}

\def\eqref#1{equation~\ref{#1}}

\def\1{\bm{1}}

\newcommand{\valid}{\mathcal{D_{\mathrm{valid}}}}

\DeclareMathAlphabet{\mathsfit}{\encodingdefault}{\sfdefault}{m}{sl}
\SetMathAlphabet{\mathsfit}{bold}{\encodingdefault}{\sfdefault}{bx}{n}

\usepackage[T1]{fontenc}   
\usepackage{hyperref}
\hypersetup{hidelinks}
\usepackage{url}
\usepackage{booktabs}
\usepackage{tabularx}
\usepackage{longtable}
\usepackage{xcolor}
\usepackage{colortbl}
\usepackage{graphicx}
\usepackage{caption}
\usepackage{placeins}
\usepackage{wrapfig}
\usepackage{needspace}
\usepackage{amssymb}
\usepackage{enumitem}
\usepackage{algorithm}
\usepackage[noend]{algpseudocode}
\usepackage{listings}
\title{MLE-World: Predicting Machine Learning Experiment Outcomes \\ with a Learned World Model}

\author{Brian Ko$^{1}$\thanks{Correspondence to: \href{mailto:kkm97183@uw.edu}{\texttt{kkm97183@uw.edu}}.}, Han Wang$^{2}$, Hejia Geng$^{3}$, Li Erran Li$^{4}$, Sherry Yang$^{5}$ \\
$^{1}$University of Washington \quad $^{2}$University of Illinois at Urbana-Champaign \\
$^{3}$University of Oxford \quad $^{4}$Columbia University \quad $^{5}$New York University}

\newif\ifshownotes
\shownotestrue
\ifshownotes
  \newcommand{\note}[2][red]{{\color{#1}\textbf{[Brian:} #2\textbf{]}}}
\else
  \newcommand{\note}[2][red]{\ignorespaces}
\fi

\iclrfinalcopy 

\newcommand{\rng}{\textendash}
\newsavebox{\wmFidelityTableBox}
\begin{document}

\maketitle
\lhead{MLE-World}

\begin{abstract}
AI research agents can propose, execute, and improve programs that solve machine learning tasks. However,
executing each candidate program requires training and evaluating ML models, limiting how many candidates an agent can explore at each step. We introduce MLE-World, a world model finetuned from Qwen3.5-4B that predicts a program's
terminal output (error traces, training logs, and validation scores) from its code and data description without executing it,
together with a confidence score estimating the prediction's correctness. Because a prediction costs one inference call rather than a training run, MLE-World
makes test time scaling (TTS) affordable for these agents. We build five TTS policies, guided by MLE-World, that compare
several candidates per step by their predicted outputs: Sequential revision, Best-of-$N$,
Pairwise judging, Fusion, and a UCB style rule that also adds an exploration bonus from this
confidence score. Since predictions can be wrong, we run a candidate for real only when its
confidence score is low. On 17 MLE-bench Lite tasks,
MLE-World matches the real terminal output on 77.9\% of held out programs, versus 41.1\% for much
larger prompted models. MLE-World guided TTS achieves up to 22.1\% higher observed mean
performance than the agent without a world model, with less real execution.
Project webpage: \url{https://mle-world.github.io/}.
\end{abstract}

\par\noindent
\begin{minipage}{\textwidth}
\centering
\includegraphics[width=\textwidth]{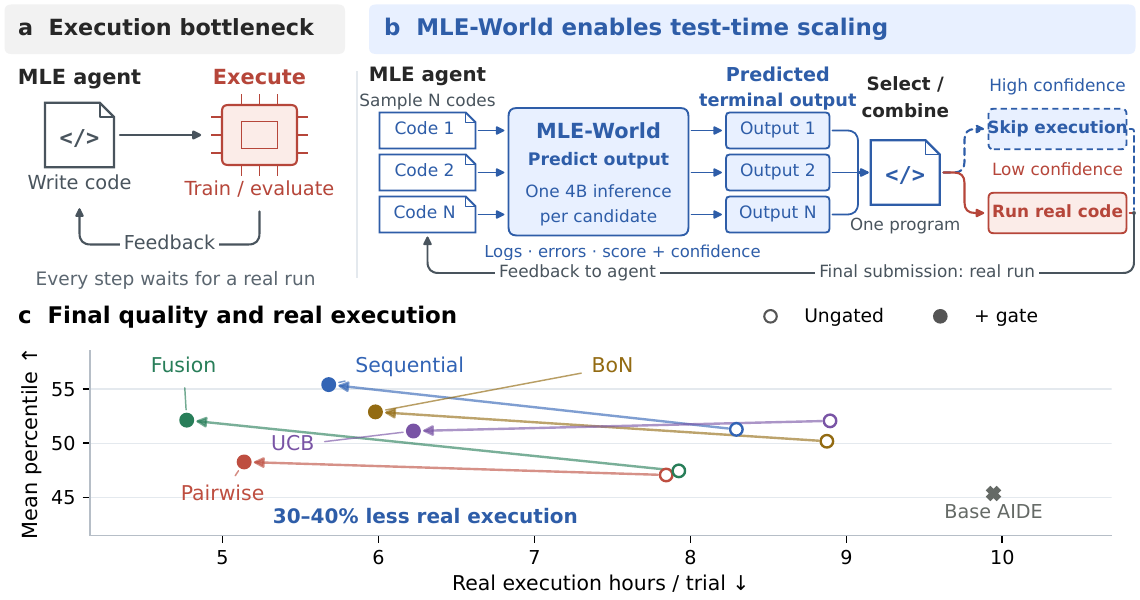}
\captionsetup{font=small,skip=4pt,hypcap=false}
\captionof{figure}{\textbf{Fast, cheap feedback with MLE-World.}
(a) Real execution. (b) Predict terminal output per candidate, select/combine, then gate.
(c) World model guided policies have higher observed mean percentiles than the agent without a world model; gating
(open to filled arrows) reduces real execution time while broadly maintaining performance. The x-axis reports mean elapsed time spent on real execution. All results are averaged over 17 tasks $\times$ three seeds
(51 trials per setting), with a 12\,h budget per trial.}
\label{fig:mle-world-overview}
\label{fig:agent-performance}
\end{minipage}
\par\clearpage

\begin{figure}[!t]
\centering
\includegraphics[width=\textwidth]{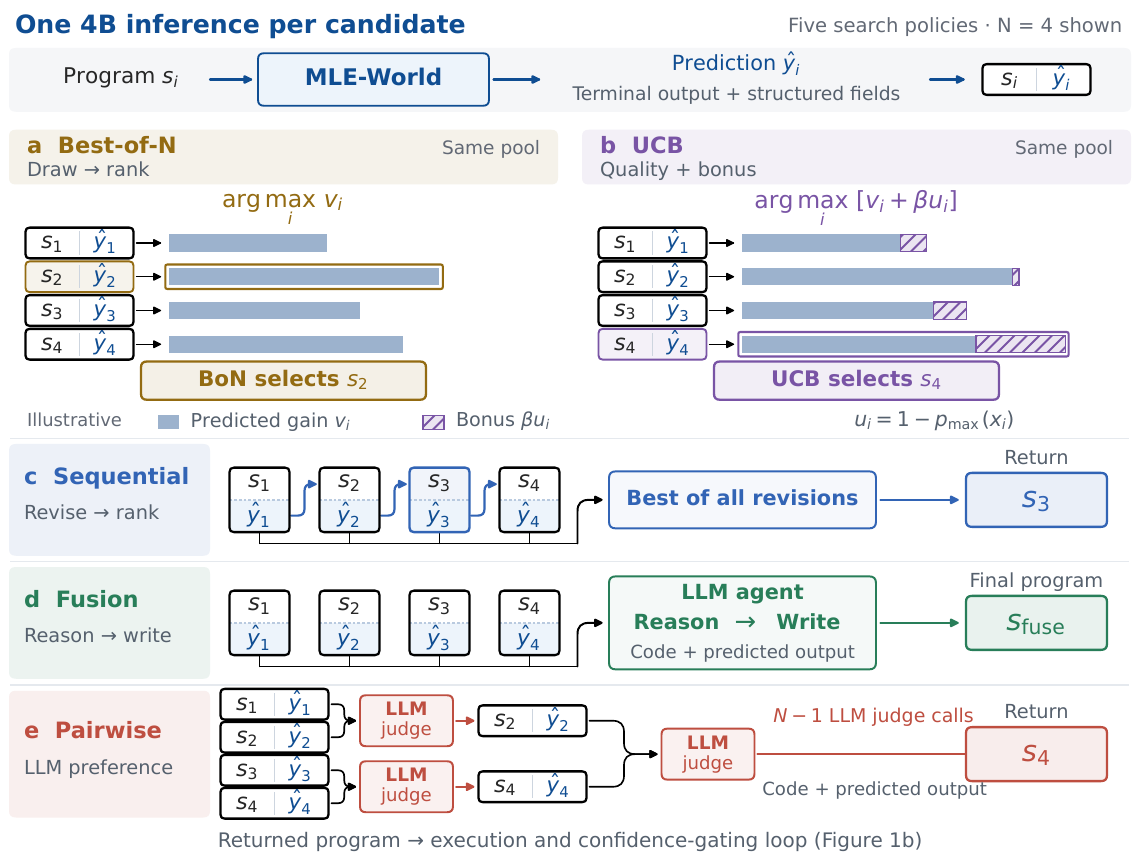}
\captionsetup{font=small}
\caption{\textbf{Five ways to search with MLE-World.}
Each tile pairs a candidate program with its predicted execution output.
Sequential revises programs using predicted feedback; Best-of-$N$ selects the candidate with the best predicted score;
Pairwise compares candidates in pairs; Fusion combines ideas into a new program;
and UCB balances predicted score with an exploration bonus for uncertain candidates.
Each policy uses four candidates and returns one program.}
\label{fig:mle-world-policies}
\end{figure}

\section{Introduction}
\label{sec:intro}

AI research agents repeatedly write programs, execute them, inspect terminal output, and revise solutions
\citep{aide,mlestar}, but this feedback is expensive because each step trains and evaluates a model.
Therefore, execution time limits how many ideas they can test
\citep{aira,aira2,rlmle}, making cheap feedback and selective execution central to efficient search.

Prior work prunes unpromising branches \citep{mlmaster,aira}, predicts model performance
\citep{kompeteai}, or ranks candidates before execution \citep{foreagent,rpm}. Pruning and ranking
can reduce total execution cost, but the cost per execution remains unchanged, and they do not provide terminal output,
such as error traces and training logs, to guide subsequent debugging and program revision.

The high cost of training makes machine learning engineering (MLE) a natural setting for world models
that predict execution feedback without running the program \citep{position}.
MLE-World predicts a candidate solution's terminal output from its code and the task's data
description. We finetune Qwen3.5-4B on logged pairs of code and output from successful and failed runs.
One inference call predicts error traces, training logs, and the validation score, supporting both
debugging and candidate selection.

Cheap predicted feedback makes test time scaling (TTS) affordable for MLE agents.
TTS can scale along two dimensions: sampling more complete trajectories \citep{monkeys,snell}
or taking more interaction steps within a trajectory \citep{tti}. In MLE, both require
additional training runs, limiting search under a fixed time budget. Prior agents execute
every candidate \citep{mlestar} or rank candidates without execution feedback \citep{foreagent}.
We focus on scaling interactions so the agent can adapt to feedback throughout the search.
MLE-World makes this affordable by letting the agent compare several candidates using
predicted error traces, training logs, and scores, then execute only the selected candidate.
We build five policies on these
predictions: Best-of-$N$ (BoN), Sequential revision, Pairwise judging, Fusion, and a rule
based on upper confidence bounds (UCB).

MLE-World also reports a confidence score with each prediction, which serves search in two ways.
First, it guides selection. Greedy selection by validation score can limit exploration
\citep{aira,aira2} and overlook candidates the world model undervalues, so UCB adds an uncertainty
bonus to predicted quality. Second, it decides when to trust a prediction. Wrong predictions can
mislead search \citep{mbpo}, so a confidence gate trusts confident predictions and runs the real execution. The gate works with any of the five policies.

Across 17 MLE-bench Lite tasks \citep{mlebench}, MLE-World matches observed terminal
output on 77.9\% of held out programs versus 41.1\% for the best prompted baseline,
with 13\rng{}1,061$\times$ less GPU inference time. MLE-World guided search
achieves up to 22.1\% higher observed mean performance than the agent without a world
model while reducing real execution. These results suggest that cheap feedback from a world
model enables test time scaling, improving agent performance by exploring more candidate programs.

We summarize our main contributions as follows:

\begin{itemize}
\item \textbf{An execution world model for MLE.} We introduce MLE-World, a 4B model that reads a program's
code and data description and predicts its terminal output, including crashes, training logs, and
validation scores, outperforming much larger prompted models.
\item \textbf{World model guided test time scaling.} We build five TTS policies that use MLE-World
to select among several candidates per step by their predicted execution feedback, then run only the
selected one. All five have higher observed mean percentiles than the agent without a world model.
\item \textbf{Selective execution using model confidence.} We run a program for real only when
MLE-World's prediction has low confidence. Gating reduces real execution time while broadly maintaining performance.
\end{itemize}
\section{Related work}
\label{sec:related}

\paragraph{MLE agents.}
MLE agents reduce execution cost through pruning
\citep{mlmaster,aira}, performance prediction \citep{kompeteai}, or candidate ranking
\citep{rpm,naspredictors}. ForeAgent \citep{foreagent} uses pairwise preference and a confidence
gate to select a candidate for execution. Scores and preferences guide selection but lack error
traces for debugging, which may require waiting for costly runs to fail. MLE-World predicts
terminal output to support both selection and revision, using confidence to decide when real
execution is needed.

\paragraph{World models for LLM agents.}
World models predict an action's outcome without performing it \citep{worldmodels,dreamer,mbpo}.
LLM world models, prompted \citep{llmpriors,bytesized32,webdreamer} or trained on agent logs
\citep{wma,webevolver,dynathink,cuwm,sweworld,agentworld}, serve web, computer use, and software
agents. Closest to ours, DSWorld \citep{dsworld} simulates individual steps in a data science
workflow and uses operation cost to decide when to simulate. MLE-World instead predicts a complete
program's terminal output, including error traces, training logs, and validation scores, to guide
program search. It uses prediction confidence to decide when real execution is needed, targeting
unreliable feedback rather than expensive operations alone.

\paragraph{Test time scaling and search.}
Sampling \citep{monkeys,snell}, tree search \citep{tot,lats}, revision \citep{reflexion,selfdebug},
synthesis \citep{fusionofn}, and longer interaction \citep{tti} require feedback to guide search.
In MLE, slow training and evaluation limit this exploration. MLE-World predicts terminal output
to guide candidate selection, revision, and synthesis, using confidence to determine when real
execution is needed.

\section{Background}
\label{sec:background}

\subsection{Agent scaffold}
\label{sec:scaffold}

\paragraph{AIDE.} We build on AIDE \citep{aide} (although MLE-World is compatible with other agent scaffolds), the reference scaffold of MLE-bench \citep{mlebench} and the base
of most later MLE agents \citep{mlmaster,aira,automind}. AIDE searches for a program $s$ that solves
task $\tau$ over a tree whose nodes are programs and whose edges are three operators: Draft writes a
new program, Debug rewrites a buggy parent, and Improve makes one atomic change to a valid parent.
For fixed execution settings, the real executor $E$ returns terminal output $o = E(\tau, s)$: training
logs, the error trace on failure, and the validation score the program computes on a held out split
of the training data. An LLM parser reads $o$ and returns a bug flag and the score. The policy drafts
a few initial solutions, debugs buggy nodes up to a fixed depth, and otherwise improves the best
valid node; the node with the best validation score is submitted and graded on the hidden test
set against the leaderboard. Every node costs one execution.

\subsection{World models for LLM agents}
\label{sec:wm-agents}

\paragraph{LLM world models.} A language world model predicts the observation resulting from an
action without executing it \citep{wma,webdreamer}. In the general formulation, an LLM world model
conditions on the observation--action history $(o_1,a_1,\ldots,o_t,a_t)$.
Because each candidate program is self-contained, MLE-World uses only the current program and
execution context to predict its terminal output, without requiring the preceding interaction history:
\[
\hat{o}_{t+1}=f_\phi(\kappa,o_t,a_t).
\]
Here $f_\phi$ is the world model with parameters $\phi$, and $\kappa$ provides the context needed
to predict the terminal output, such as the environment state. We treat the complete candidate
program as the current observation $o_t$ and its execution as the action $a_t$. The model predicts
the resulting terminal output as the next observation $\hat{o}_{t+1}$.
Section~\ref{sec:method} details the model's input and structured output format.

\section{MLE-World}
\label{sec:method}

MLE-World implements this prediction with a language model $h_\phi$ whose input $x$ explicitly
encodes the program and execution context. Before execution, it predicts a richer record $y$ that
augments terminal output $o_{t+1}$ with three structured fields:
\begin{equation}
\label{eq:record}
\begin{aligned}
x &= \big\langle\, \underset{\text{program}}{o_t},\ \ \underset{\text{packages}}{P_{\mathrm{env}}},\ \ \underset{\text{data overview}}{D_\tau},\ \ \underset{\text{time limit}}{T} \,\big\rangle,\\
y &= \big\langle\, \underset{\text{exception type}}{c},\ \ \underset{\text{terminal output}}{o_{t+1}},\ \ \underset{\text{validation metric}}{m},\ \ \underset{\text{submission flag}}{b} \,\big\rangle,
\end{aligned}
\end{equation}
Here we write $s=o_t$ for the program and $o=o_{t+1}$ for its terminal output, matching
Section~\ref{sec:scaffold}, and use $s$ and $o$ below. The fields of $y$ are decoded in the order shown, with
$c \in \mathcal{C} = \{\texttt{null}, \texttt{ValueError}, \texttt{ImportError}, \dots\}$ (all 82 classes in Appendix~\ref{app:exception-classes}).
Each exception label is one added token, decoded first. Hats denote predictions:
$\hat y=h_\phi(x)=\langle\hat c,\hat o,\hat m,\hat b\rangle$; $\hat o$ is the predicted terminal output.
We write $x_s$ for the full input associated with program $s$, and $x_i=x_{s_i}$ for candidate $i$.
We finetune
Qwen3.5-4B with standard supervised finetuning (SFT) on logged pairs $(x, y) \in \mathcal{D}$, one
execution per sequence, minimizing the negative log likelihood of the output tokens,
\begin{equation}
\label{eq:loss}
\mathcal{L}(\phi) = -\,\mathbb{E}_{(x, y) \sim \mathcal{D}} \big[ \log p_\phi(y \mid x) \big]
\end{equation}
where $p_\phi(y\mid x)$ is the conditional output distribution defined by $h_\phi$
(training configuration in Appendix~\ref{app:training}). At search time, these predictions drive candidate selection and
the Debug and Improve operators, and the gate of Section~\ref{sec:execution-gate} decides when to
obtain real feedback instead.

\label{sec:input-target}

\paragraph{Input.} Outcomes depend on the data, available libraries, and execution budget as well
as the code. MLE-World receives the program $s$, installed packages and versions $P_{\mathrm{env}}$,
AIDE's data overview $D_\tau$ (a file tree summary, CSV dimensions, and column summaries with
example values), and time limit $T$.

\paragraph{Target.} Predicting the terminal output alone is not enough. Without an explicit
exception type, the model must infer from free text whether the run crashed, and AIDE reads the
validation score only through a separate feedback model that extracts it from the output
(Section~\ref{sec:scaffold}). We therefore predict both as their own fields, whose labels AIDE
already logs. Adding them raises agreement on terminal output from 59.4\% to 62.8\% and metric recall
from 84.3\% to 89.5\% (Appendix~\ref{app:ablation}). The exception type also yields the outcome
distribution behind our confidence score, defined next.

\label{sec:uncertainty-score}
\paragraph{Confidence score from the first output token.}
The first decoded token is the predicted exception class $\hat c \in \mathcal{C}$ of Equation~(\ref{eq:record}), which
routes search to Debug or Improve. Its token probabilities, renormalized over $\mathcal C$, define the
outcome distribution $p_\phi(c\mid x)$ and the scores
\begin{equation}
\label{eq:uncertainty}
p_{\max}(x) = \max_{c \in \mathcal{C}} p_\phi(c \mid x),
\qquad u(x) = 1 - p_{\max}(x),
\qquad p_{\mathrm{clean}}(x) = p_\phi(c = \texttt{null} \mid x).
\end{equation}
$p_{\max}$ is the confidence in the most likely outcome, the uncertainty score $u$ is the mass on all
others, and $p_{\mathrm{clean}}$ is the probability of a clean run, equal to $p_{\max}$ when a clean run is
most likely. Across quintiles of $p_{\max}$ on held out programs, every predicted field matches the
real run more often as $p_{\max}$ rises (Figure~\ref{fig:wm-confidence-fidelity}), so $p_{\max}$ is a
suitable confidence score. Search uses this outcome distribution twice: UCB (Section~\ref{sec:ucb-selection}) uses
$p_{\mathrm{clean}}$ to score candidates, and the gate (Section~\ref{sec:execution-gate}) uses
$p_{\max}$ to decide whether the chosen one must run.

\label{sec:data}
\paragraph{Data.} Every executed AIDE node yields one training pair $(x, y)$. We run AIDE with five agent models from
4B to 122B parameters on the 17 MLE-bench tasks and keep failed episodes, since a world model
trained only on clean runs is least accurate on buggy candidates, where the search needs it most,
and mixing agent policies makes a simulator robust to the agent that later uses it
\citep{word2world}. Appendix~\ref{app:train-data} reports statistics of the training data (Table~\ref{tab:train-stats}).

\section{Search guided by MLE-World}
\label{sec:uncertainty-search}

We build five world model guided TTS policies on MLE-World's predictions (Section~\ref{sec:wm-policies}), then add a
confidence gate that works with any of them and decides when a prediction needs a real run
(Section~\ref{sec:execution-gate}).

\subsection{Test time scaling on predicted outputs}
\label{sec:wm-policies}

\paragraph{Predicted outputs let a step screen many candidates and run one.}
MLE-World stands in for execution when comparing candidates: a step draws $N$ candidates, predicts their outputs
$\hat y_i = h_\phi(x_i)$, and executes only one. Each of our five policies reads these
predictions, not only the code (Figure~\ref{fig:mle-world-policies}; algorithms in
Appendix~\ref{app:tts-algorithms}). \emph{Best-of-$N$ (BoN)} keeps the best predicted validation score
\citep{monkeys,mlestar}. \emph{Sequential} revises each candidate from the previous one's predicted
error trace, log, and score, where prior revision uses real feedback \citep{snell,selfdebug,mlestar}.
\emph{Pairwise} extends ForeAgent \citep{foreagent}, which compares candidates without execution
feedback: an LLM judge compares two candidates at a time with their predicted terminal outputs until one remains.
\emph{Fusion} synthesizes a new program from the pool and its predicted terminal outputs
\citep{fusionofn,mlestar}. \emph{UCB} adds an exploration bonus from MLE-World's confidence to the
predicted score, as described next.

\label{sec:ucb-selection}
\paragraph{UCB: greedy selection can discard promising programs.}
BoN selects the best predicted validation score without accounting for prediction reliability.
If MLE-World underestimates a promising program's quality, greedy selection can discard it before
execution provides feedback for subsequent search.

\paragraph{Outcome uncertainty supplies an inexpensive exploration signal.}
GP-UCB balances estimated value with the objective's posterior standard deviation \mbox{\citep{gpucb}}.
We use outcome uncertainty $u(x)$ as an empirical exploration proxy, not a formal confidence bound.
Lower confidence accompanies less accurate metric predictions on held out programs
(Figure~\ref{fig:wm-confidence-fidelity}), so the bonus favors potentially underestimated candidates
at no additional inference cost.

\paragraph{The bonus balances predicted quality and exploration.}
At search step $t$ the agent draws $N$ candidates; $\mathcal{S}_t$ holds those predicted to complete
normally, write a submission, and have a finite predicted metric, using the same pool as BoN. We write $\hat m(s_i)=\hat m_i$
for the metric field of $\hat y_i$. Greedy selection on this pool chooses $\arg\max_{s\in\mathcal{S}_t} d_\tau\hat{m}(s)$
($d_\tau=+1$ for a maximized metric and $-1$ for a minimized metric). An uncertain prediction may underestimate the true
quality, so adding the bonus gives such a candidate a chance to beat a confidently predicted one with a
slightly better estimate:
\begin{equation}
\label{eq:ucb}
s_t^\star = \operatorname*{arg\,max}_{s \in \mathcal{S}_t} \bigl[v(s) + \beta u(x_s)\bigr],
\qquad
v(s) = \frac{d_\tau\bigl(\hat{m}(s) - m_{\mathrm{ref}}\bigr)}{\alpha_t},
\end{equation}
where $s_t^\star$ is selected and $v(s)$ measures predicted gain over $m_{\mathrm{ref}}$ in units of
$\alpha_t>0$. Both terms are set at step $t$ and shared within its pool; with a verified reference,
$\alpha_t=\max(|m_{\mathrm{ref}}|,0.01)$ (fallbacks in Appendix~\ref{app:tts-algorithms}).
The UCB runs decode greedily, making $\texttt{null}$ the most probable class on $\mathcal S_t$;
therefore $u(x_s)=1-p_{\max}(x_s)=1-p_{\mathrm{clean}}(x_s)$ there. The weight $\beta\geq0$ controls
exploration, and the gate (Section~\ref{sec:execution-gate}) runs low confidence selections for real.
Without an eligible metric, UCB, like BoN, returns the first
usable program or asks an LLM judge (Appendix~\ref{app:tts-algorithms}).

\subsection{Execution gating by uncertainty}
\label{sec:execution-gate}

\paragraph{Real runs are spent where predictions are least reliable.}
Predictions are cheap but not always right: on held out programs, 22\% of predicted terminal outputs
and 14\% of predicted exception classes disagree with the real run (Table~\ref{tab:wm-fidelity}), and
such errors can mislead search \citep{mbpo}. Because accuracy rises with confidence
(Section~\ref{sec:uncertainty-score}), after any of the five policies selects a candidate, the gate
requests real feedback when
\begin{equation}
\label{eq:execution-gate}
p_{\max}(x_{s_t^\star}) < \gamma,
\end{equation}
or when confidence is missing, where $s_t^\star$ is the program returned by the search policy
at step $t$ and $\gamma\in(0,1)$ is a confidence threshold. Otherwise the predicted
output guides the next Debug or Improve step. We always execute a candidate with a new best
predicted validation metric, regardless of confidence, because the best verified candidate supplies
the final submission.

\paragraph{Thresholds follow the cost of each predicted outcome.}
We set $\gamma=\gamma_{\mathrm{s}}$ for predicted successes and $\gamma=\gamma_{\mathrm{f}}$ for failures, because their
runs cost very different amounts of time. A failing program usually crashes within seconds, so we can afford to run many of them, and each run returns the exact error trace to repair.
A successful program trains to completion, which is where execution time goes, so
a lower $\gamma_{\mathrm{s}}$ skips more of these long runs. Programs predicted to time out are never run, since a timeout is the most costly outcome.

\section{Experiments}
\label{sec:experiments}

The experiments answer three questions:
\textbf{(Q1)} How accurately does MLE-World predict execution outcomes, and at what inference cost?
(Section~\ref{sec:fidelity})
\textbf{(Q2)} Does world model guided search improve submitted solutions, and do predicted outputs
and the exploration bonus each help?
(Section~\ref{sec:search-results})
\textbf{(Q3)} Does confidence identify unreliable predictions, and what tradeoff between quality and
execution cost does gating achieve? (Section~\ref{sec:gating-results})
Our main experiments use the same 17 of the 22 MLE-bench Lite tasks \citep{mlebench}: following
\citet{rlmle}, we exclude five tasks on which our agents rarely make progress
(task selection and statistics of the training data in Appendix~\ref{app:train-data}).
Following MLE-bench \citep{mlebench}, we select the submission with the best real validation score
in each trial for grading on the hidden test set.

\subsection{MLE-World predicts execution accurately at low cost}
\label{sec:fidelity}

\begin{figure}[t]
\centering
\captionsetup{font=small,skip=5pt}
\captionof{table}{\textbf{MLE-World leads all prediction fields.} Accuracy (\%) on 1,700 held out programs;
$\pm$ SE from 2,000 program bootstrap resamples. Exception class: exact match; terminal output and
crash class: judged match; metric: within 5\% relative error. Corpus baselines: Appendix~\ref{app:inference}.}
\label{tab:wm-fidelity}
\sbox{\wmFidelityTableBox}{%
\small
\setlength{\tabcolsep}{2.4pt}%
\begingroup

\begin{tabular}{@{}lrrrr@{}}
\toprule
 & \begin{tabular}[b]{@{}r@{}}Terminal\\output\end{tabular} & \begin{tabular}[b]{@{}r@{}}Exception\\class\end{tabular} & \begin{tabular}[b]{@{}r@{}}Crash\\class\end{tabular} & \begin{tabular}[b]{@{}r@{}}Validation\\metric\end{tabular} \\
\midrule
\multicolumn{5}{@{}l}{\emph{Prompted models}} \\
DeepSeek & 41.1{\scriptsize$\pm$1.1} & 57.2{\scriptsize$\pm$1.2} & 48.2{\scriptsize$\pm$1.5} & 18.6{\scriptsize$\pm$1.7} \\
Coder & 17.8{\scriptsize$\pm$0.9} & 27.5{\scriptsize$\pm$1.1} & 13.2{\scriptsize$\pm$1.0} & 16.3{\scriptsize$\pm$1.7} \\
\multicolumn{5}{@{}l}{\emph{Corpus, no model}} \\
Task majority & -- & 34.2{\scriptsize$\pm$1.2} & -- & 55.2{\scriptsize$\pm$2.3} \\
Nearest program & 28.0{\scriptsize$\pm$1.1} & 41.1{\scriptsize$\pm$1.2} & 14.0{\scriptsize$\pm$1.0} & 55.6{\scriptsize$\pm$2.2} \\
Parent record & 22.3{\scriptsize$\pm$1.0} & 34.2{\scriptsize$\pm$1.2} & 6.7{\scriptsize$\pm$0.7} & 48.8{\scriptsize$\pm$2.3} \\
\midrule
\textbf{MLE-World} & \textbf{77.9}{\scriptsize$\pm$1.0} & \textbf{86.4}{\scriptsize$\pm$0.8} & \textbf{76.4}{\scriptsize$\pm$1.3} & \textbf{78.1}{\scriptsize$\pm$1.9} \\
\bottomrule
\end{tabular}
\endgroup
}
\begin{minipage}[c]{0.60\textwidth}
\centering
\usebox{\wmFidelityTableBox}
\end{minipage}\hfill
\begin{minipage}[c]{0.38\textwidth}
\centering
\includegraphics[width=\linewidth,
height=\dimexpr\ht\wmFidelityTableBox+\dp\wmFidelityTableBox\relax,
keepaspectratio,trim={9.625bp 10.3bp 2.875bp 2.2bp},clip]{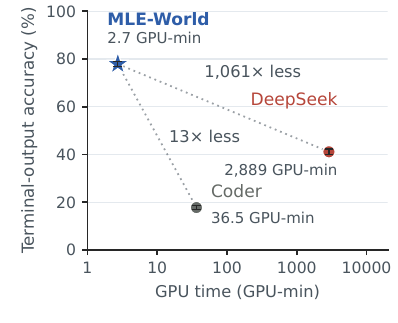}
\end{minipage}
\captionof{figure}{\textbf{Accurate prediction at low inference cost.} Terminal output accuracy versus
GPU time (wall clock $\times$ GPUs; log scale).}
\label{fig:wm-fidelity-vs-time}
\end{figure}

We answer \textbf{(Q1)} by evaluating MLE-World in isolation on a disjoint held out set:
per field accuracy of the predicted output and inference cost, against much larger prompted models.

\paragraph{Training setup.}
We finetune Qwen3.5-4B on the 497,056 training programs of
Section~\ref{sec:method}, with its input and target of four fields (hyperparameters and hardware in Appendix~\ref{app:training}).
Adding the exception class and validation metric to the target improves terminal output agreement,
so we use this four field target (ablation in Appendix~\ref{app:ablation}).

\paragraph{Evaluation setup.}
We evaluate on 1,700 held out programs, 100 per task and balanced across agent stages. No held out
program has an exact or near duplicate copy in training (Appendix~\ref{app:dedup} describes the
deduplication), but the tasks are shared by design. As in model based RL, where a world model predicts new
actions in the environment it was trained on \citep{worldmodels,dreamer,mbpo}, the task is the
environment and each new program is an action. We compare against two strong prompted coder models, DeepSeek-V4-Flash-0731 (DeepSeek) and
Qwen3-Coder-Next (Coder), and three corpus baselines that
copy records from MLE-World's training data without a model: task majority, nearest program, and
parent record (see Appendix~\ref{app:inference} for details of each method). Because exact
matching misses rewordings and conflates crashes with different causes, a Coder judge scores terminal
output and crash class; it agrees with held out human labels on 96.7\% (Appendix~\ref{app:judge}).

\paragraph{1) MLE-World predicts outcomes beyond task priors and record retrieval.}
Table~\ref{tab:wm-fidelity} shows that MLE-World's predicted terminal output matches the observed
output on 77.9\% of held out programs, compared with 41.1\% for DeepSeek, the strongest prompted baseline. The corpus controls clarify what the model learns: task level
statistics already predict many validation scores, but retrieving a similar program's record
rarely reproduces its failure cause, since a one line difference in code can produce a different error. MLE-World's advantage on both fields supports learning
relationships between programs and their execution outcomes beyond those captured by the tested
baselines. As shown in Figure~\ref{fig:wm-fidelity-vs-time}, its GPU inference time is
13\rng{}1,061$\times$ lower than that of the prompted models, making these predictions practical for screening multiple candidates.

\paragraph{2) MLE-World misses what reading the program cannot reveal.}
Table~\ref{tab:wm-breakdown} shows that MLE-World leads both prompted models at every stage and
outcome. Comparing it with Table~\ref{tab:train-stats}, its errors follow what the input can show, not
how often an outcome appears in training. A \texttt{SyntaxError} is visible in the code, so MLE-World predicts
every one, although they are only 8.2\% of training crashes.
Timeouts are its weakest
outcome because run time must be estimated from model size, epochs, and data size.

\subsection{Observed solution quality with world model guided search}
\label{sec:search-results}
\label{sec:medals}

We answer \textbf{(Q2)} in three steps: full search runs without confidence gating, single decisions
on fixed task states, and UCB's exploration bonus in isolation.

\begin{figure}[!t]
\centering
\begin{minipage}[c]{0.49\textwidth}
\centering
\resizebox{\linewidth}{!}{
\begingroup
\fontsize{8.5}{10.2}\selectfont
\setlength{\tabcolsep}{3pt}
\renewcommand{\arraystretch}{1.25}
\begin{tabular}{@{}lrrr@{}}
\toprule
Policy & \shortstack[r]{Mean\\percentile $\uparrow$} & \shortstack[r]{$\Delta$ vs.\\Base} & \shortstack[r]{Valid\\(\%) $\uparrow$} \\
\midrule
Base AIDE & 45.4{\scriptsize$\pm$4.8} & -- & 84.3{\scriptsize$\pm$2.0} \\
\midrule
\multicolumn{4}{@{}l}{\emph{MLE-World guided (12 h)}} \\
\rowcolor[gray]{0.95}
Sequential & 51.3{\scriptsize$\pm$5.7} & +5.9{\scriptsize$\pm$4.1} & \textbf{100.0}{\scriptsize$\pm$0.0} \\
Pairwise & 47.1{\scriptsize$\pm$2.8} & +1.7{\scriptsize$\pm$2.5} & 90.2{\scriptsize$\pm$5.2} \\
\rowcolor[gray]{0.95}
Fusion & 47.5{\scriptsize$\pm$3.3} & +2.1{\scriptsize$\pm$3.1} & \textbf{100.0}{\scriptsize$\pm$0.0} \\
BoN & 50.2{\scriptsize$\pm$1.8} & +4.8{\scriptsize$\pm$3.3} & 94.1{\scriptsize$\pm$0.0} \\
\rowcolor[gray]{0.95}
UCB & \textbf{52.1}{\scriptsize$\pm$0.8} & \textbf{+6.7}{\scriptsize$\pm$2.5} & 98.0{\scriptsize$\pm$2.0} \\
\bottomrule
\end{tabular}
\endgroup
}
\end{minipage}\hfill
\begin{minipage}[c]{0.49\textwidth}
\centering
\includegraphics[width=\linewidth]{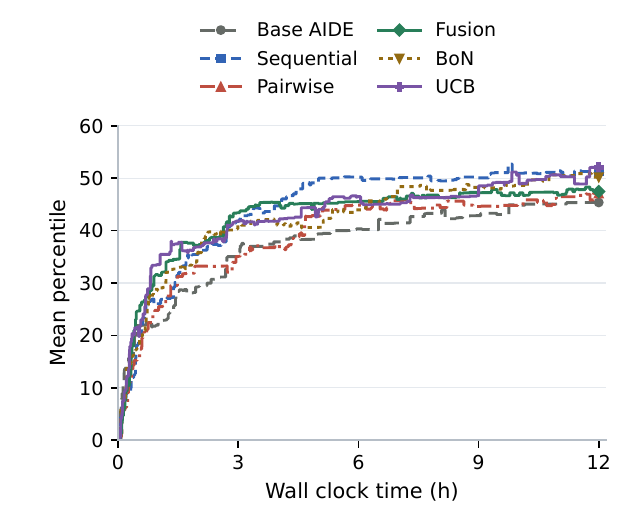}
\end{minipage}
\par\smallskip
\begin{minipage}[t]{0.49\textwidth}
\captionsetup{font=small,skip=5pt}
\captionof{table}{\textbf{Search without confidence gating.}
Mean $\pm$ SE across seed means; $\Delta$: gain over Base AIDE $\pm$ paired task bootstrap SE.
Bold: best 12\,h mean.}
\label{tab:ungated-agent-performance}
\end{minipage}\hfill
\begin{minipage}[t]{0.49\textwidth}
\captionsetup{font=small,skip=5pt}
\captionof{figure}{\textbf{Search quality over time.} Percentile of the validation-selected submission,
carried forward and zero until valid.}
\label{fig:search-wall-clock}
\end{minipage}
\end{figure}

\paragraph{Setup.}
We use Qwen3.5-9B to write code and Qwen3-Coder-Next as the feedback model that provides code reviews
within the AIDE scaffold. We run each setting on 17 tasks with three seeds per task and report results
averaged across tasks and seeds. We select $\beta=0.5$ for UCB through
tuning (Appendix~\ref{app:ucb-beta}). We report mean leaderboard percentile because medal thresholds amplify
noise and hide progress on hard tasks (\citealp{aira2}; Appendix~\ref{app:medal-metric}).

\paragraph{1) All five policies have higher observed means than the agent without a world model.}
UCB reaches 52.1 mean percentile versus 45.4 for Base AIDE\footnote{In a separate reference, real-execution BoN runs all four
candidates and reaches 41.9 at 8\,h (Appendix~\ref{app:agent-execution}).}; all guided policies have
higher observed means (Table~\ref{tab:ungated-agent-performance}). They also spend less time on
crashes and timeouts (Figure~\ref{fig:execution-split}), consistent with predictions steering
execution toward programs that run and score.

\par\noindent
\begin{minipage}[t]{0.56\textwidth}
\vspace{0pt}
\paragraph{2) Predicted terminal output helps search policies.}
To isolate the benefit of predicted terminal output, we compare each policy with and without it
from the same starting states. On 150 shared states,
Pairwise and Fusion receive the same four input candidates; Sequential makes three revisions from
the first. Fusion adds one synthesis call before retries; Pairwise uses three judge calls. With predicted
terminal output, Fusion has a higher observed mean than with code alone:
20.6 versus 14.0 (Table~\ref{tab:fixed-candidates-compact}). This suggests that predicted
terminal output improves search by helping policies select or synthesize better candidates
within each step.
\end{minipage}\hfill
\begin{minipage}[t]{0.41\textwidth}
\vspace{0pt}
\centering
\begingroup
\fontsize{9}{10.8}\selectfont
\setlength{\tabcolsep}{3pt}
\renewcommand{\arraystretch}{1.08}
\begin{tabularx}{\linewidth}{@{}Xrr@{}}
\toprule
Policy & \shortstack[r]{w/o predicted\\terminal output} & \shortstack[r]{w/ predicted\\terminal output} \\
\midrule
Pairwise & 13.0 & 15.4 \\
Sequential & 10.4 & 15.3 \\
Fusion & 14.0 & \textbf{20.6} \\
\bottomrule
\end{tabularx}
\endgroup

\captionsetup{font=small,skip=5pt,hypcap=false}
\captionof{table}{\textbf{Value of predicted terminal output.} Mean percentile on 150 shared states, 17 tasks;
invalid outputs score zero. Sequential always selects by predicted metric.
Details: Table~\ref{tab:fixed-candidates}.}
\label{tab:fixed-candidates-compact}
\end{minipage}
\par
\begin{figure}[!htb]
\centering
\begin{minipage}[t]{0.60\textwidth}
\vspace{0pt}
\paragraph{3) The exploration bonus yields a higher observed mean.}
To isolate the exploration bonus, we vary only Improve-stage $\beta$: 0 (greedy) versus 1 (UCB),
with other settings fixed (Appendix~\ref{app:ucb-improve-control}). We focus on Improve because most
Draft and Debug candidates are buggy and have no validation metric.
With the bonus, the observed mean percentile is 45.4,
versus 38.6 for greedy selection (Figure~\ref{fig:ucb-improve-main}). This suggests that the bonus
can help search by favoring candidates whose predictions are uncertain.
\end{minipage}\hfill
\begin{minipage}[t]{0.37\textwidth}
\vspace{0pt}
\centering
\includegraphics[width=\linewidth]{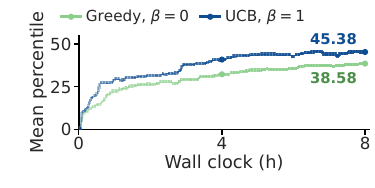}
\captionsetup{font=footnotesize,skip=3pt,hypcap=false}
\captionof{figure}{\textbf{The exploration bonus yields a higher observed mean.} All other settings are
fixed; only Improve-stage $\beta$ varies (Figure~\ref{fig:ucb-ablations}a).}
\label{fig:ucb-improve-main}
\end{minipage}
\end{figure}
\par

\begin{figure}[!t]
\centering
\begin{minipage}[c]{0.49\textwidth}
\centering
\resizebox{\linewidth}{!}{
\begingroup
\fontsize{8.5}{10.2}\selectfont
\setlength{\tabcolsep}{2.5pt}
\renewcommand{\arraystretch}{1.18}
\begin{tabular}{@{}lrrrr@{}}
\toprule
\multicolumn{5}{@{}l}{\emph{Mean percentile $\uparrow$}} \\
Policy & Ungated & Gated & $\Delta$ & \\
\midrule
Base AIDE & 45.4 & -- & -- & \\
\rowcolor[gray]{0.95}
Sequential & 51.3 & 55.4 & +4.1{\scriptsize$\pm$6.4} & \\
Pairwise & 47.1 & 48.3 & +1.2{\scriptsize$\pm$4.1} & \\
\rowcolor[gray]{0.95}
Fusion & 47.5 & 52.1 & +4.7{\scriptsize$\pm$1.3} & \\
BoN & 50.2 & 52.9 & +2.7{\scriptsize$\pm$2.1} & \\
\rowcolor[gray]{0.95}
UCB & 52.1 & 51.1 & -0.9{\scriptsize$\pm$2.4} & \\
\midrule
\multicolumn{5}{@{}l}{\emph{Real execution (h/trial) $\downarrow$}} \\
Policy & Ungated & Gated & $\Delta$ & Saved (\%) \\
\midrule
Base AIDE & 9.94 & -- & -- & -- \\
\rowcolor[gray]{0.95}
Sequential & 8.29 & 5.68 & -2.61{\scriptsize$\pm$0.55} & 31.5 \\
Pairwise & 7.84 & 5.14 & -2.71{\scriptsize$\pm$0.34} & 34.5 \\
\rowcolor[gray]{0.95}
Fusion & 7.93 & 4.77 & -3.16{\scriptsize$\pm$0.37} & 39.8 \\
BoN & 8.87 & 5.98 & -2.89{\scriptsize$\pm$0.47} & 32.6 \\
\rowcolor[gray]{0.95}
UCB & 8.90 & 6.22 & -2.67{\scriptsize$\pm$0.48} & 30.0 \\
\bottomrule
\end{tabular}
\endgroup
}
\end{minipage}\hfill
\begin{minipage}[c]{0.49\textwidth}
\centering
\includegraphics[width=\linewidth]{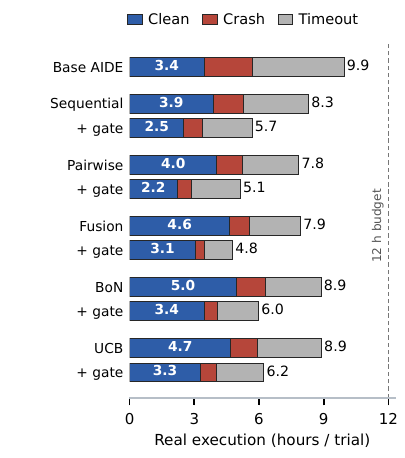}
\end{minipage}
\par\smallskip
\begin{minipage}[t]{0.49\textwidth}
\captionsetup{font=small,skip=5pt,hypcap=false}
\captionof{table}{\textbf{Gated versus ungated performance and real execution time.}
$\Delta$: gated minus ungated, $\pm$ SE across the three paired seed means.
Wall clock curves: Figure~\ref{fig:gated-wall-clock} (Appendix~\ref{app:agent-execution}).}
\label{tab:confidence-gating}
\end{minipage}\hfill
\begin{minipage}[t]{0.49\textwidth}
\captionsetup{font=small,skip=5pt,hypcap=false}
\captionof{figure}{\textbf{Execution time by code output.} World model guidance reduces time spent on
crashes and timeouts, while confidence gating further reduces real execution time.}
\label{fig:execution-split}
\end{minipage}
\end{figure}

\subsection{Confidence gating: execution savings and observed quality}
\label{sec:gating-results}

We answer \textbf{(Q3)} by additionally running each selection policy with the confidence gate enabled.

\begin{samepage}
\paragraph{Gating setup.}
As defined in Section~\ref{sec:execution-gate}, $\gamma_{\mathrm{s}}$ and $\gamma_{\mathrm{f}}$ are
confidence thresholds for predicted successes and failures, respectively. We set $\gamma_{\mathrm{s}}=0.50$
for all five policies and $\gamma_{\mathrm{f}}=0.56$ for Sequential, Pairwise, and BoN,
and $0.83$ for Fusion and UCB. These threshold values are selected from a three-hour tuning sweep
(Appendix~\ref{app:gate-thresholds}). All other settings match the ungated runs.
\end{samepage}

\begin{figure}[!ht]
\centering
\includegraphics[width=0.88\textwidth]{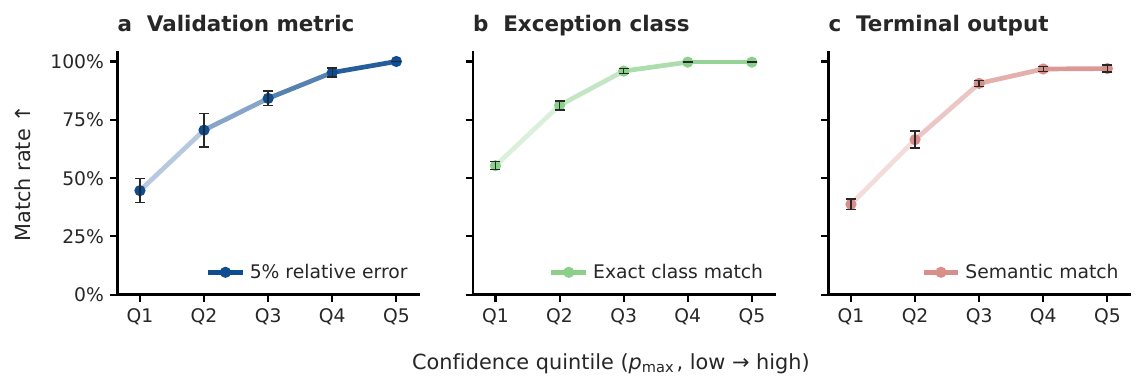}
\captionsetup{font=small,skip=5pt}
\caption{\textbf{Prediction fidelity rises with confidence.} Accuracy by $p_{\max}$ quintile on 1,700 held out
programs: (a) metric within 5\% relative error,
(b) exact exception class, (c) judged terminal output. Bars: $\pm$ task bootstrap SE.}
\label{fig:wm-confidence-fidelity}
\end{figure}

\paragraph{1) Prediction fidelity rises steadily with confidence.}
We group the held out programs from Section~\ref{sec:fidelity} into quintiles of $p_{\max}$.
Figure~\ref{fig:wm-confidence-fidelity} shows that match rates for validation metrics, exception
classes, and terminal outputs rise with confidence. This suggests spending real execution on low-confidence
predictions, where errors are more likely, while relying on predicted feedback when confidence is high.

\FloatBarrier

\paragraph{2) Confidence gating reduces real execution time while broadly maintaining performance.}
Table~\ref{tab:confidence-gating} shows that across the five policies, gating saves 30\rng{}40\% of
real execution time while broadly maintaining performance (trajectories in
Figure~\ref{fig:gated-wall-clock}). Gated Sequential achieves the highest
observed mean percentile. The savings can buy more search steps within the same budget
(Appendix~\ref{app:agent-execution}).

\paragraph{3) Confidence contributes beyond random skipping.}
In the separate 8\,h control in Table~\ref{tab:random-gate} (Appendix~\ref{app:random-gate}), for BoN, confidence gating
uses 19\% less real execution than a random gate that skips runs with a fixed probability, at a
similar final percentile. This suggests that the savings come from which runs the gate skips, not
only how many.

\section{Limitations and conclusion}
\label{sec:limitations}

\paragraph{Limitations.}
Our main evaluation is in distribution: new programs on 17 tasks seen in finetuning
(Appendix~\ref{app:dedup}). On three tasks excluded from finetuning, the 4B MLE-World, without
reasoning, still predicts terminal output more accurately than much larger prompted models (53.3\%
versus 42.1\% for DeepSeek with reasoning and 21.1\% for Coder; Appendix~\ref{app:ood}). This opens
a generalizable world model as future work, for example by adding reasoning or scaling up the model.
Training uses about 500,000 prior executions, including earlier trajectories from a 9B agent but no
reported runs, favoring settings with accumulated logs. Zero shot baselines and Base AIDE lack this task knowledge;
nonparametric corpus baselines trail MLE-World (Table~\ref{tab:wm-fidelity}), while Base AIDE with corpus access
remains untested. Real execution determines final scores; prediction errors can misdirect search. Timeouts are
the least accurately predicted outcome, and a missed timeout still costs a full run; improving runtime prediction is future work.

\paragraph{Conclusion.}
A 4B model finetuned on logged executions predicts the outcomes of new programs more accurately
than much larger prompted models, at 13\rng{}1,061$\times$ less GPU inference time. This makes test
time scaling affordable for MLE agents: all five world model guided policies have higher observed mean
percentiles than the agent without a world model. Gating reduces real execution time by 30\rng{}40\% while broadly maintaining performance.

\section*{Ethics statement}
This work raises no ethical concerns. Its experiments use only public Kaggle competition data
distributed with MLE-bench and involve no human subjects or personal data.

\section*{Reproducibility statement}
Appendices~\ref{app:train-data} to~\ref{app:parsing} give the training data, training configuration,
deduplication and holdout, decoding and the prediction prompt, and output parsing; Appendices~\ref{app:judge},
\ref{app:tts-algorithms}, \ref{app:gate-thresholds}, and~\ref{app:random-gate} give the judge prompt
and validation, search algorithms, gate thresholds, and random control. Agent results state seeds,
budgets, and evaluation protocols. MLE-World was trained on 8 NVIDIA H200 GPUs
(full training configuration in Appendix~\ref{app:training}). We will release
the MLE-World checkpoint, the execution corpus with its held out split, all prompts, and the agent and
evaluation code.

\section*{LLM usage}
LLMs are part of the method: Qwen3.5 models generate agent programs and Qwen3-Coder-Next
judges outputs. An LLM coding agent (GPT-5.6 Sol) developed the judge prompt
under the protocol of Appendix~\ref{app:judge}. The authors also used LLM assistants to write
experiment and analysis code and to edit the text, and remain responsible for all content.

\clearpage
\bibliography{iclr2027_conference}
\bibliographystyle{iclr2027_conference}

\appendix
\makeatletter
\setlength{\@fptop}{0pt}
\setlength{\@fpsep}{12pt}
\setlength{\@fpbot}{0pt plus 1fil}
\makeatother

\section{Training data}
\label{app:train-data}

Table~\ref{tab:train-stats} breaks the training set down by generator model, agent stage, outcome,
and task. Every task in the holdout is also a training task; the held out programs themselves are
excluded (Appendix~\ref{app:dedup}).

\paragraph{Task selection.} MLE-bench Lite contains 22 tasks. Following \citet{rlmle}, who evaluate
on the tasks where their base model could generate a valid initial solution, we keep the 17 tasks
on which our agents produce working programs often enough to train on and to search. On the four
excluded image tasks (\texttt{aptos2019-\allowbreak blindness-\allowbreak detection},
\texttt{histopathologic-\allowbreak cancer-\allowbreak detection},
\texttt{ranzcr-\allowbreak clip-\allowbreak catheter-\allowbreak line-\allowbreak classification},
and \texttt{siim-\allowbreak isic-\allowbreak melanoma-\allowbreak classification}), only
2.3\rng{}11.6\% of deduplicated executions across all agent models finish cleanly with a submission,
against 38.6\% on the 17 kept tasks; our corpus contains no executions of
the fifth,
\texttt{the-\allowbreak icml-\allowbreak 2013-\allowbreak whale-\allowbreak challenge-\allowbreak right-\allowbreak whale-\allowbreak redux}.

\begin{longtable}{@{}lrr@{}}
\caption{\textbf{Training data statistics.} Most training programs fail: 52.3\% crash and 7.0\%
time out, so MLE-World sees failures more often than clean runs. Counts over the 497,056 training
programs, each recovered to its task, generator model, and stage from the deduplicated node pool.
Outcome, exception class, and labels come from the training target; token counts use the
Qwen3.5-4B tokenizer, and the prompt includes the system message.}
\label{tab:train-stats} \\
\toprule
 & Programs & \% \\
\midrule
\endfirsthead
\caption[]{Training data statistics (continued).} \\
\toprule
\endhead
All training programs & 497,056 & 100.0 \\
\midrule
\multicolumn{3}{@{}l}{\emph{Generator model}} \\
\quad Qwen3-Coder-Next & 209,573 & 42.2 \\
\quad Qwen3.5-4B & 129,907 & 26.1 \\
\quad Qwen3.5-122B-A10B & 77,239 & 15.5 \\
\quad Qwen3.5-9B & 54,815 & 11.0 \\
\quad Qwen3.6-35B-A3B & 25,522 & 5.1 \\
\midrule
\multicolumn{3}{@{}l}{\emph{Agent stage}} \\
\quad Draft & 106,542 & 21.4 \\
\quad Improve & 183,875 & 37.0 \\
\quad Debug & 206,639 & 41.6 \\
\midrule
\multicolumn{3}{@{}l}{\emph{Outcome}} \\
\quad Clean run & 202,408 & 40.7 \\
\quad Crash & 259,884 & 52.3 \\
\quad Timeout / interrupt & 34,764 & 7.0 \\
\midrule
\multicolumn{3}{@{}l}{\emph{Crash exception class (\% of crashes)}} \\
\quad ValueError & 72,882 & 28.0 \\
\quad TypeError & 36,718 & 14.1 \\
\quad NameError & 25,578 & 9.8 \\
\quad KeyError & 21,987 & 8.5 \\
\quad SyntaxError & 21,380 & 8.2 \\
\quad AttributeError & 20,846 & 8.0 \\
\quad RuntimeError & 15,362 & 5.9 \\
\quad FileNotFoundError & 10,352 & 4.0 \\
\quad Other & 34,779 & 13.4 \\
\midrule
\multicolumn{3}{@{}l}{\emph{Labels}} \\
\quad Numeric validation metric & 193,562 & 38.9 \\
\quad Submission written & 199,081 & 40.1 \\
\midrule
Task & Programs & \\
\midrule
aerial-cactus-identification & 22,705 & \\
denoising-dirty-documents & 26,357 & \\
detecting-insults-in-social-commentary & 31,255 & \\
dog-breed-identification & 30,278 & \\
dogs-vs-cats-redux-kernels-edition & 53,010 & \\
jigsaw-toxic-comment-classification-challenge & 15,724 & \\
leaf-classification & 16,603 & \\
mlsp-2013-birds & 57,706 & \\
new-york-city-taxi-fare-prediction & 15,589 & \\
nomad2018-predict-transparent-conductors & 37,299 & \\
plant-pathology-2020-fgvc7 & 27,075 & \\
random-acts-of-pizza & 45,494 & \\
spooky-author-identification & 21,896 & \\
tabular-playground-series-dec-2021 & 28,218 & \\
tabular-playground-series-may-2022 & 20,160 & \\
text-normalization-challenge-english-language & 26,783 & \\
text-normalization-challenge-russian-language & 20,904 & \\
\midrule
\multicolumn{3}{@{}l}{\emph{Length, median (90th pct.)}} \\
Prompt tokens & 3,421 (5,406) & \\
Target tokens & 186 (617) & \\
Program lines & 203 (369) & \\
\bottomrule
\end{longtable}

\section{Training configuration}
\label{app:training}

We train all parameters for five epochs with LlamaFactory \citep{llamafactory} on 8 H200 GPUs
(43 hours) and use the final checkpoint. The input is the program with its importable packages and
versions, the task's data overview, and the execution time limit, which lets the model predict
timeouts. Table~\ref{tab:training-config} lists the LlamaFactory configuration of the MLE-World checkpoint
evaluated in Section~\ref{sec:experiments}. Each exception class is a single added token
(\texttt{EXC\_ValueError}, \dots), so the class is decoded as one token rather than spelled out.
Holdout loss was logged during training but not used to select the checkpoint.

\begin{table}[h]
\centering
\caption{\textbf{Training configuration.} SFT of all parameters of Qwen3.5-4B; we evaluate the final
checkpoint (step 6,685).}
\label{tab:training-config}

\begin{tabular}{@{}ll@{}}
\toprule
Base model & Qwen3.5-4B \\
Method & SFT of all parameters (LlamaFactory), vision tower frozen \\
Training data & 497,056 programs (Table~\ref{tab:train-stats}) \\
Chat template & \texttt{qwen3\_5}, thinking disabled \\
Added tokens & 82 exception class tokens (\texttt{EXC\_None}, \texttt{EXC\_ValueError}, \dots; Appendix~\ref{app:exception-classes}) \\
Sequence length & 12,288 tokens, packed without attention across examples \\
Epochs / steps & 5 / 6,685 \\
Batch size & 128 packed sequences (8 GPUs $\times$ 1 $\times$ 16 accumulation) \\
Optimizer & AdamW (fused), weight decay 0.01, gradient clipping 1.0 \\
Learning rate & $1\times10^{-5}$, cosine schedule, 5\% warmup \\
Precision & bf16 mixed precision, FP32 weights and optimizer states \\
Memory & Gradient checkpointing, FlashAttention-2, Liger kernels \\
Hardware & 8 NVIDIA H200 GPUs (4 nodes $\times$ 2), 42.7 hours \\
Final loss & 0.079 train, 0.081 holdout \\
\bottomrule
\end{tabular}
\end{table}

\section{Exception classes}
\label{app:exception-classes}

Table~\ref{tab:exception-classes} lists the set $\mathcal{C}$ of Equation~(\ref{eq:record}): the 82
exception classes MLE-World can predict, each added to the tokenizer as one token
(\texttt{EXC\_}\emph{name}). The set covers every exception class that occurs in the training corpus,
named as the logged traceback reports it; \texttt{None} is the no exception class, written
\texttt{null} in the main text. Because each class is a single token, the first decoded token gives
a distribution over all 82 classes, from which Equation~(\ref{eq:uncertainty}) reads
$p_{\max}$ and $p_{\mathrm{clean}}$.

\begin{table}[h]
\centering
\caption{\textbf{Exception classes.} The 82 classes of $\mathcal{C}$, in token order; each is one added
token \texttt{EXC\_}\emph{name}.}
\label{tab:exception-classes}

\begin{tabular}{@{}ll@{}}
\toprule
\texttt{None} (\texttt{null}) & \texttt{MissingCorpusError} \\
\texttt{AcceleratorError} & \texttt{ModuleNotFoundError} \\
\texttt{ArrowInvalid} & \texttt{NameError} \\
\texttt{ArrowTypeError} & \texttt{NotADirectoryError} \\
\texttt{AssertionError} & \texttt{NotFittedError} \\
\texttt{AttributeError} & \texttt{NotImplementedError} \\
\texttt{AxisError} & \texttt{OSError} \\
\texttt{BadZipFile} & \texttt{OptionError} \\
\texttt{BdbQuit} & \texttt{OutOfBoundsDatetime} \\
\texttt{BracketError} & \texttt{OutOfMemoryError} \\
\texttt{CalledProcessError} & \texttt{OverflowError} \\
\texttt{CatBoostError} & \texttt{ParserError} \\
\texttt{CompressionError} & \texttt{PicklingError} \\
\texttt{DTypePromotionError} & \texttt{ReadError} \\
\texttt{DateParseError} & \texttt{RecursionError} \\
\texttt{DeferredCudaCallError} & \texttt{RepositoryNotFoundError} \\
\texttt{EOFError} & \texttt{RuntimeError} \\
\texttt{EmptyDataError} & \texttt{SameFileError} \\
\texttt{Error} & \texttt{SpecificationError} \\
\texttt{Exception} & \texttt{StopIteration} \\
\texttt{FileExistsError} & \texttt{SyntaxError} \\
\texttt{FileNotFoundError} & \texttt{SystemExit} \\
\texttt{FloatingPointError} & \texttt{TabError} \\
\texttt{HTTPError} & \texttt{TerminatedWorkerError} \\
\texttt{ImportError} & \texttt{TimeoutError} \\
\texttt{IndentationError} & \texttt{TorchRuntimeError} \\
\texttt{IndexError} & \texttt{TqdmKeyError} \\
\texttt{IndexingError} & \texttt{TypeError} \\
\texttt{IntCastingNaNError} & \texttt{UFuncTypeError} \\
\texttt{InvalidIndexError} & \texttt{URLError} \\
\texttt{InvalidParameterError} & \texttt{UnboundLocalError} \\
\texttt{InvalidURL} & \texttt{UnicodeDecodeError} \\
\texttt{InvalidVersion} & \texttt{UnicodeEncodeError} \\
\texttt{IsADirectoryError} & \texttt{UnidentifiedImageError} \\
\texttt{JSONDecodeError} & \texttt{UnpicklingError} \\
\texttt{KeyError} & \texttt{UnsupportedOperation} \\
\texttt{LightGBMError} & \texttt{ValueError} \\
\texttt{LinAlgError} & \texttt{VersionError} \\
\texttt{LookupError} & \texttt{XGBoostError} \\
\texttt{MemoryError} & \texttt{ZeroDivisionError} \\
\texttt{MergeError} & \texttt{error} \\
\bottomrule
\end{tabular}
\end{table}

\section{Target ablation}
\label{app:ablation}

MLE-World predicts an exception class and a validation metric alongside the terminal log
(Section~\ref{sec:method}). This ablation asks whether these fields add information or merely
duplicate what the log already contains.

\paragraph{Setup.}
We fix the backbone, training corpus, split, and decoding, and vary only the target. Starting from
terminal output plus a submission flag, we add the exception class and then the validation metric.
We also compare untrimmed logs with the default window of the first and last 2,500 characters, and we train
a target of metric plus submission flag with no terminal output. To keep each run affordable, the variants train on a subset of 49,791 programs drawn from nine of the tasks and are
scored against executed ground truth on 450 held out programs from those nine tasks
(Table~\ref{tab:target-ablation}). Absolute scores are therefore not comparable with Table~\ref{tab:wm-fidelity}.

\paragraph{Little observed difference from trimming.}
We observed little difference between the 2,500 character window and untrimmed logs
(terminal output 59.4\% vs.\ 59.5\%, exception class 71.7\% vs.\ 71.4\%;
Table~\ref{tab:target-ablation}). With one training run per condition, these point estimates
do not establish equivalence.

\paragraph{Structured fields improve the log itself.}
Adding the exception class and then the validation metric raises agreement on terminal output from
59.4\% to 62.8\%, exception class agreement from 71.7\% to 75.3\%, and crash class agreement from
64.8\% to 69.1\%. Predicting explicit outcome fields also makes the generated log more accurate,
rather than serving only as a separate output.

\paragraph{Dropping the log loses coverage and feedback.}
A target of the metric alone reaches the highest metric precision (90.0\%) but predicts a metric for only
77.0\% of cases with a numeric reference, compared with 89.5\% for the full target. It also provides no
error trace for debugging. The full target gives up 1.0 point of precision for 12.5 points of
recall and the feedback from logs the agent needs to repair failures.

\begin{table}[t]
\centering
\caption{\textbf{Target ablation.} Structured fields improve the predicted log, and dropping the log
costs metric recall. 450 held out programs from its nine training tasks; backbone, data, and
decoding are fixed. ``+'' adds fields cumulatively; trimmed logs retain the first and last 2,500
characters. All scores are percentages. Terminal output measures semantic agreement;
exception/crash measure exact class agreement inferred from logs, with ``no exception''
included and crash restricted to jointly crashing cases. Metric precision uses 10\%
relative tolerance on commonly answered cases; recall is the fraction of cases with a numeric reference
receiving a metric. Dashes indicate undefined scores from logs. Each row is one training run;
at 450 programs, a rate near 60\rng{}75\% has a bootstrap SE over programs of about 2 points (more in the crash
and metric columns, which use subsets).}
\label{tab:target-ablation}

\begin{tabularx}{\linewidth}{@{}Xrrrr@{}}
\toprule
target & \begin{tabular}[b]{@{}r@{}}Terminal\\output\end{tabular}
       & \begin{tabular}[b]{@{}r@{}}Exception\\class\end{tabular}
       & \begin{tabular}[b]{@{}r@{}}Crash\\class\end{tabular}
       & \begin{tabular}[b]{@{}r@{}}Validation metric\\(precision / recall)\end{tabular} \\
\midrule
terminal output (untrimmed) + submission flag & 59.5 & 71.4 & 64.4 & 87.3 / 83.5 \\
terminal output (trimmed) + submission flag   & 59.4 & 71.7 & 64.8 & 88.7 / 84.3 \\
\quad + exception class                       & 60.7 & 73.0 & 66.7 & 87.3 / 87.6 \\
\quad\textbf{+ validation metric}             & \textbf{62.8} & \textbf{75.3} & \textbf{69.1} & 89.0 / \textbf{89.5} \\
\midrule
validation metric + submission flag           & -- & -- & -- & \textbf{90.0} / 77.0 \\
\bottomrule
\end{tabularx}
\end{table}

\section{Deduplication and holdout construction}
\label{app:dedup}

\paragraph{Deduplication.} Agents rarely resubmit byte identical code; repeated attempts usually
differ only in comments or formatting. We therefore key each program on a hash of its Python
token stream with comments and blank lines removed and indentation and line breaks normalized.
Programs that differ only in comments, spacing, or blank lines share a key; any change to
identifiers, literals, or structure does not. Programs that fail to tokenize fall back to a hash of
their raw bytes. Before deduplication, we drop executions whose outcome reflects the harness rather
than the program, such as GPU contention, host memory limits, or network failures. Deduplication
then collapses 757,186 executions to 602,741. When the same program produced different outcomes
across runs (for example, a timeout in one run and a clean finish in another), we keep one
representative per outcome rather than an arbitrary winner.

\paragraph{Filtering.} From the 602,741 deduplicated executions we remove 43,836 on tasks outside
our 17, 33,591 whose output contains a progress bar (agents are told not to print them), 13,121
whose validation metric cannot be resolved unambiguously, 11,549 cut short by the agent's overall
deadline and logged as timeouts, and 1,888 with harness failures, invalid stage labels, or
mislabeled interrupts. This leaves 498,756 executions.

\paragraph{Holdout.} From these 498,756 executions, we sample 100 programs per task, balanced across
agent stages (draft, improve, debug) and between fast ($\le$5\,s) and slow executions. We draw only
programs without a detected exact or near duplicate among the other programs of their task
(procedure below), then remove the sampled programs from training. The remaining 497,056
executions form the training set.

\paragraph{Near duplicates.} Exact keys miss programs that differ only in identifier names or in a
few edits. We therefore replace identifiers with placeholders and treat two programs as duplicates
when their identifier normalized abstract syntax trees hash identically, and as near duplicates
when MinHash with locality sensitive hashing estimates a Jaccard
similarity of at least 0.9 between their identifier normalized token shingles (128 permutations in
16 bands of 8 rows, within each task). No exact or near duplicates were detected between holdout
and training under this procedure. The probabilistic MinHash/LSH search can miss similar pairs;
it does not rule out semantic similarity or overlap across tasks.

\paragraph{Shared tasks.} Held out programs are new, but their tasks also appear in training. This
mirrors model based RL, where a world model is trained and evaluated in the same environment and
must predict the outcome of new actions, not new environments \citep{worldmodels,dreamer,mbpo}.
Here the task is the environment and executing a program is an action: an agent writes hundreds of
new programs for the task it is solving, so MLE-World must predict unseen programs within known
tasks.

\section{Inference and judge settings}
\label{app:inference}

Table~\ref{tab:inference-settings} lists the decoding settings used to predict the holdout.
MLE-World and Coder decode greedily; DeepSeek samples one output per program at temperature 1.0
and top-$p$ 0.95. MLE-World runs without reasoning. Coder is a
nonreasoning model by default; we include it as a strong coding model, since predicting execution
requires reading code. DeepSeek runs with reasoning enabled at maximum effort, and only its final answer,
not its reasoning trace, is scored.

\begin{table}[h]
\centering
\caption{\textbf{Decoding settings.} Settings used to predict the 1,700 held out programs, and for
the judge that scores terminal output and crash class.}
\label{tab:inference-settings}

\begin{tabular}{@{}lcccc@{}}
\toprule
 & MLE-World & Coder & DeepSeek & Judge (Coder) \\
\midrule
Reasoning & off & off & on (max effort) & off \\
Temperature & 0 (greedy) & 0 (greedy) & 1.0 & 1.0 \\
Top-$p$ & 1.0 & 1.0 & 0.95 & 0.95 \\
Top-$k$ & -- & -- & -- & 40 \\
Samples per program & 1 & 1 & 1 & 3 \\
Max output tokens & 8,192 & 8,192 & 32,768 & 3,072 \\
\bottomrule
\end{tabular}
\end{table}

The judge is Coder and sees only the predicted and observed terminal outputs, not the
exception labels. Coder is also a baseline. It is the least accurate model on the judged fields of Table~\ref{tab:wm-fidelity}, so a judge bias
toward its own outputs would not favor MLE-World, but we do not rule such a bias out.

\paragraph{Prediction prompt.} All three models receive the same system and user messages, reproduced
below. The prompted baselines are zero shot, with no examples from the corpus. MLE-World emits the
exception class as one added token; the prompted models write the same \texttt{EXC\_} label as text.

\begin{lstlisting}[literate={…}{{\ldots}}1 {—}{{---}}1 {→}{{$\rightarrow$}}1]
--- system message ---
Predict the outcome and terminal output (stdout+stderr) of a Python ML script.

Input, in order:
- <code>, the script, run as runfile.py.
- <environment>, the importable packages. Authoritative: a library not listed is not importable, and the listed version is the one whose API applies.
- <data_overview>, truth for files/columns under ./input/.
- <runtime_budget>, the wall-clock limit. The script is killed when it is exceeded.

Output, in order and nothing else:
- <exc_type>…</exc_type>, one EXC_<ClassName> token: EXC_None if it runs clean, otherwise its exception class prefixed with EXC_ (e.g. EXC_ValueError); EXC_TimeoutError if its wall-clock time would exceed <runtime_budget>.
- <term_out>…</term_out>, the script's stdout+stderr. Write it before the two fields below — they are read off it.
- <val_metric>…</val_metric>, the validation score reported in the term_out, or None if it reports none, the validation is invalid (e.g. scoring training data or computing the wrong metric), or the script raises.
- <submission_exists>true|false</submission_exists>, whether ./submission/submission.csv exists after the run.

Missing files or columns → predict the right Python exception, don't invent data.
A script can finish without raising and still write no submission — decide <submission_exists> from what the code actually writes to ./submission/, not from whether it crashed.

--- user message ---
<code>
{program, run as runfile.py}
</code>

<environment>
{Python version; importable packages with versions; packages not installed or failing to import}
</environment>

<data_overview>
{file tree under ./input/ with sizes, plus columns and sample rows, as AIDE shows the agent}
</data_overview>

<runtime_budget>{execution time limit, e.g. 3400s}</runtime_budget>
\end{lstlisting}

\paragraph{Corpus baselines.}
These baselines use no model. They see the same 497,056 training programs as MLE-World, or the
parent's observed record, and copy records verbatim in MLE-World's target format, so the parser,
judge, and bootstrap of Table~\ref{tab:wm-fidelity} apply unchanged (Table~\ref{tab:corpus-baselines}).
\emph{Task majority} predicts, for each task and stage, the most common exception class and
submission flag of the training programs and the median of their validation metrics; it writes no
terminal output. \emph{Nearest program} copies the record of the training program of the same task
whose code is most similar (TF-IDF cosine over code tokens and token bigrams; median cosine 0.70).
\emph{Parent record} copies the observed record of the program's parent, which an agent already holds
when it writes an Improve or Debug child; the 561 drafts and 14 programs whose parent is missing from
the logs fall back to the nearest program. The nearest training program is the parent itself for 26\%
of Improve and 48\% of Debug programs. \emph{Random record} copies a random training program of the
same task and stage.

\begin{table}[h]
\centering
\caption{\textbf{Corpus baselines.} Baselines built from MLE-World's training corpus without a
model trail MLE-World on every field and stage. Accuracy (\%, $\pm$ SE from 2,000 bootstrap resamples over programs)
on the 1,700 held out programs; stage columns give Draft/Improve/Debug. Parent record equals Nearest
program on the 561 drafts; their draft terminal output scores differ by 0.1 only because each
baseline was judged separately by the sampled judge.}
\label{tab:corpus-baselines}

\begin{tabular}{@{}lrrrr@{}}
\toprule
Method & \shortstack[r]{Terminal\\output} & \shortstack[r]{Exception\\class} & \shortstack[r]{Crash\\class} & \shortstack[r]{Validation\\metric} \\
\midrule
Task majority & -- & 34.2{$\,\pm$1.2} & -- & 55.2{$\,\pm$2.3} \\
Nearest program & 28.0{$\,\pm$1.1} & 41.1{$\,\pm$1.2} & 14.0{$\,\pm$1.0} & 55.6{$\,\pm$2.2} \\
Parent record & 22.3{$\,\pm$1.0} & 34.2{$\,\pm$1.2} & 6.7{$\,\pm$0.7} & 48.8{$\,\pm$2.3} \\
Random record & 11.2{$\,\pm$0.8} & 22.2{$\,\pm$1.0} & 3.5{$\,\pm$0.6} & 20.2{$\,\pm$1.8} \\
\midrule
MLE-World & 77.9{$\,\pm$1.0} & 86.4{$\,\pm$0.8} & 76.4{$\,\pm$1.3} & 78.1{$\,\pm$1.9} \\
\bottomrule
\end{tabular}
\par\medskip
\begin{tabular}{@{}lrr@{}}
\toprule
Method & \shortstack[r]{Terminal output\\D/I/Dbg} & \shortstack[r]{Exception class\\D/I/Dbg} \\
\midrule
Task majority & -- & 34.4/43.5/25.1 \\
Nearest program & 23.2/44.3/16.8 & 39.4/55.1/29.1 \\
Parent record & 23.3/35.1/8.8 & 39.4/43.5/20.2 \\
Random record & 11.3/17.5/5.1 & 20.9/31.4/14.7 \\
\midrule
MLE-World & 78.3/83.2/72.3 & 86.5/89.8/82.9 \\
\bottomrule
\end{tabular}

\end{table}

\section{Output parsing}
\label{app:parsing}

Each prediction must return the exception class, terminal output, and validation metric in tagged
fields. A missing field receives no credit in Table~\ref{tab:wm-fidelity}, except a missing
exception class, which is read from the predicted terminal output.
Table~\ref{tab:parse-rates} reports how often each model returned each field.

\begin{table}[h]
\centering
\caption{\textbf{Output parse rates.} Prompted models rarely fail to parse, so parsing does not
explain their low accuracy. Share (\%) of the 1,700 held out predictions that finish before the token cap and
contain each scored field; a missing exception field is read from the terminal output. Both
prompted models return all three fields for over 91\% of programs, far too few failures to
account for their 28\rng{}60 percentage point gap in Table~\ref{tab:wm-fidelity}. DeepSeek's output is long
because it reasons, while the other two models run with reasoning off.}
\label{tab:parse-rates}

\begin{tabular}{@{}lrrr@{}}
\toprule
 & DeepSeek & Coder & \shortstack[r]{MLE-World\\(4B + SFT)} \\
\midrule
Finished before token cap & 94.9 & 91.8 & 99.7 \\
\midrule
Exception class & 93.6 & 100.0 & 100.0 \\
Terminal output & 91.8 & 91.8 & 99.7 \\
Validation metric & 91.8 & 91.8 & 99.7 \\
All three fields & 91.4 & 91.8 & 99.7 \\
\midrule
Median output tokens & 12,042 & 144 & 160 \\
\bottomrule
\end{tabular}

\end{table}

\section{Accuracy breakdown}
\label{app:breakdown}

Table~\ref{tab:wm-breakdown} splits the accuracy of Table~\ref{tab:wm-fidelity} by agent stage,
observed outcome, and crash type.

\begin{table}[!htbp]
\centering
\caption{\textbf{Accuracy by stage, outcome, and crash type.} MLE-World leads both prompted
baselines on every slice and field except the crash class of \texttt{IndexError} (26 programs),
and timeouts are its hardest case. Accuracy (\%) of MLE-World (MLE), DeepSeek (DS), and
Coder; $n$ counts programs per row. Darker cells are more accurate and the best value per field is bold. \emph{Crash type} splits the 1,094 crashing programs by their observed exception class; the
ten most frequent are listed. Crash class is scored only on program errors and the metric only on
programs with a numeric reference, so some cells are empty.}
\label{tab:wm-breakdown}

\definecolor{wmacc}{HTML}{2C7FB8}
\begin{tabular}{@{}lrrrrrrr@{}}
\toprule
 & & \multicolumn{3}{c}{Terminal output} & \multicolumn{3}{c}{Exception class} \\
\cmidrule(lr){3-5}\cmidrule(l){6-8}
 & $n$ & MLE & DS & Coder & MLE & DS & Coder \\
\midrule
\multicolumn{8}{@{}l}{\emph{Stage}} \\
\quad Draft & 561 & \cellcolor{wmacc!35}\textbf{78.3} & \cellcolor{wmacc!18}40.9 & \cellcolor{wmacc!7}16.5 & \cellcolor{wmacc!39}\textbf{86.5} & \cellcolor{wmacc!25}55.4 & \cellcolor{wmacc!12}26.6 \\
\quad Improve & 561 & \cellcolor{wmacc!37}\textbf{83.2} & \cellcolor{wmacc!18}38.9 & \cellcolor{wmacc!10}21.4 & \cellcolor{wmacc!40}\textbf{89.8} & \cellcolor{wmacc!26}57.2 & \cellcolor{wmacc!13}29.8 \\
\quad Debug & 578 & \cellcolor{wmacc!33}\textbf{72.3} & \cellcolor{wmacc!20}43.5 & \cellcolor{wmacc!7}15.5 & \cellcolor{wmacc!37}\textbf{82.9} & \cellcolor{wmacc!27}59.0 & \cellcolor{wmacc!12}26.3 \\
\midrule
\multicolumn{8}{@{}l}{\emph{Outcome}} \\
\quad Clean run & 525 & \cellcolor{wmacc!39}\textbf{86.8} & \cellcolor{wmacc!16}35.6 & \cellcolor{wmacc!14}30.4 & \cellcolor{wmacc!43}\textbf{95.6} & \cellcolor{wmacc!27}59.6 & \cellcolor{wmacc!19}43.2 \\
\quad Crash & 1,094 & \cellcolor{wmacc!34}\textbf{76.0} & \cellcolor{wmacc!21}46.7 & \cellcolor{wmacc!6}12.9 & \cellcolor{wmacc!38}\textbf{83.7} & \cellcolor{wmacc!26}58.8 & \cellcolor{wmacc!9}20.2 \\
\quad Timeout / interrupt & 81 & \cellcolor{wmacc!21}\textbf{46.1} & \cellcolor{wmacc!1}1.6 & \cellcolor{wmacc!1}1.6 & \cellcolor{wmacc!28}\textbf{61.7} & \cellcolor{wmacc!9}21.0 & \cellcolor{wmacc!11}24.7 \\
\midrule
\multicolumn{8}{@{}l}{\emph{Crash type}} \\
\quad ValueError & 230 & \cellcolor{wmacc!30}\textbf{67.2} & \cellcolor{wmacc!19}41.4 & \cellcolor{wmacc!1}2.3 & \cellcolor{wmacc!35}\textbf{78.3} & \cellcolor{wmacc!24}53.0 & \cellcolor{wmacc!2}4.3 \\
\quad SyntaxError & 161 & \cellcolor{wmacc!41}\textbf{90.1} & \cellcolor{wmacc!16}34.8 & \cellcolor{wmacc!5}12.0 & \cellcolor{wmacc!45}\textbf{100.0} & \cellcolor{wmacc!30}66.5 & \cellcolor{wmacc!13}29.2 \\
\quad TypeError & 133 & \cellcolor{wmacc!32}\textbf{70.9} & \cellcolor{wmacc!17}36.8 & \cellcolor{wmacc!1}2.0 & \cellcolor{wmacc!35}\textbf{76.7} & \cellcolor{wmacc!22}49.6 & \cellcolor{wmacc!0}0.8 \\
\quad NameError & 115 & \cellcolor{wmacc!43}\textbf{96.2} & \cellcolor{wmacc!38}84.9 & \cellcolor{wmacc!26}57.4 & \cellcolor{wmacc!44}\textbf{98.3} & \cellcolor{wmacc!41}90.4 & \cellcolor{wmacc!35}77.4 \\
\quad AttributeError & 104 & \cellcolor{wmacc!34}\textbf{75.3} & \cellcolor{wmacc!22}48.1 & \cellcolor{wmacc!4}9.6 & \cellcolor{wmacc!36}\textbf{79.8} & \cellcolor{wmacc!23}51.9 & \cellcolor{wmacc!3}7.7 \\
\quad KeyError & 86 & \cellcolor{wmacc!31}\textbf{69.0} & \cellcolor{wmacc!23}51.9 & \cellcolor{wmacc!5}12.0 & \cellcolor{wmacc!36}\textbf{80.2} & \cellcolor{wmacc!29}64.0 & \cellcolor{wmacc!12}26.7 \\
\quad RuntimeError & 53 & \cellcolor{wmacc!26}\textbf{58.5} & \cellcolor{wmacc!7}15.7 & \cellcolor{wmacc!0}0.6 & \cellcolor{wmacc!35}\textbf{77.4} & \cellcolor{wmacc!17}37.7 & \cellcolor{wmacc!3}5.7 \\
\quad FileNotFoundError & 43 & \cellcolor{wmacc!34}\textbf{76.0} & \cellcolor{wmacc!28}62.8 & \cellcolor{wmacc!7}14.7 & \cellcolor{wmacc!38}\textbf{83.7} & \cellcolor{wmacc!32}72.1 & \cellcolor{wmacc!19}41.9 \\
\quad ImportError & 38 & \cellcolor{wmacc!40}\textbf{89.5} & \cellcolor{wmacc!27}60.5 & \cellcolor{wmacc!4}8.8 & \cellcolor{wmacc!40}\textbf{89.5} & \cellcolor{wmacc!27}60.5 & \cellcolor{wmacc!1}2.6 \\
\quad IndexError & 26 & \cellcolor{wmacc!18}\textbf{41.0} & \cellcolor{wmacc!16}35.9 & \cellcolor{wmacc!2}3.8 & \cellcolor{wmacc!22}\textbf{50.0} & \cellcolor{wmacc!21}46.2 & \cellcolor{wmacc!12}26.9 \\
\quad Other & 105 & \cellcolor{wmacc!35}\textbf{76.8} & \cellcolor{wmacc!22}47.9 & \cellcolor{wmacc!7}15.6 & \cellcolor{wmacc!36}\textbf{80.0} & \cellcolor{wmacc!21}46.7 & \cellcolor{wmacc!6}13.3 \\
\bottomrule
\end{tabular}
\par\medskip
\begin{tabular}{@{}lrrrrrrr@{}}
\toprule
 & & \multicolumn{3}{c}{Crash class} & \multicolumn{3}{c}{Validation metric} \\
\cmidrule(lr){3-5}\cmidrule(l){6-8}
 & $n$ & MLE & DS & Coder & MLE & DS & Coder \\
\midrule
\multicolumn{8}{@{}l}{\emph{Stage}} \\
\quad Draft & 561 & \cellcolor{wmacc!35}\textbf{78.2} & \cellcolor{wmacc!21}46.8 & \cellcolor{wmacc!5}11.0 & \cellcolor{wmacc!33}\textbf{74.0} & \cellcolor{wmacc!8}18.1 & \cellcolor{wmacc!6}13.4 \\
\quad Improve & 561 & \cellcolor{wmacc!36}\textbf{79.1} & \cellcolor{wmacc!20}44.8 & \cellcolor{wmacc!7}15.7 & \cellcolor{wmacc!39}\textbf{85.7} & \cellcolor{wmacc!9}20.7 & \cellcolor{wmacc!9}19.4 \\
\quad Debug & 578 & \cellcolor{wmacc!33}\textbf{72.7} & \cellcolor{wmacc!23}52.0 & \cellcolor{wmacc!6}13.4 & \cellcolor{wmacc!30}\textbf{67.5} & \cellcolor{wmacc!7}15.0 & \cellcolor{wmacc!6}13.3 \\
\midrule
\multicolumn{8}{@{}l}{\emph{Outcome}} \\
\quad Clean run & 525 & -- & -- & -- & \cellcolor{wmacc!35}\textbf{78.1} & \cellcolor{wmacc!8}18.6 & \cellcolor{wmacc!7}16.3 \\
\quad Crash & 1,094 & \cellcolor{wmacc!34}\textbf{76.4} & \cellcolor{wmacc!22}48.2 & \cellcolor{wmacc!6}13.2 & -- & -- & -- \\
\quad Timeout / interrupt & 81 & -- & -- & -- & -- & -- & -- \\
\midrule
\multicolumn{8}{@{}l}{\emph{Crash type}} \\
\quad ValueError & 230 & \cellcolor{wmacc!30}\textbf{67.4} & \cellcolor{wmacc!20}43.6 & \cellcolor{wmacc!1}2.2 & -- & -- & -- \\
\quad SyntaxError & 161 & \cellcolor{wmacc!41}\textbf{90.1} & \cellcolor{wmacc!16}34.8 & \cellcolor{wmacc!5}12.0 & -- & -- & -- \\
\quad TypeError & 133 & \cellcolor{wmacc!32}\textbf{71.2} & \cellcolor{wmacc!17}38.3 & \cellcolor{wmacc!1}1.8 & -- & -- & -- \\
\quad NameError & 115 & \cellcolor{wmacc!43}\textbf{96.5} & \cellcolor{wmacc!39}86.1 & \cellcolor{wmacc!27}59.4 & -- & -- & -- \\
\quad AttributeError & 104 & \cellcolor{wmacc!34}\textbf{75.6} & \cellcolor{wmacc!22}49.0 & \cellcolor{wmacc!4}9.6 & -- & -- & -- \\
\quad KeyError & 86 & \cellcolor{wmacc!32}\textbf{71.3} & \cellcolor{wmacc!26}57.0 & \cellcolor{wmacc!6}13.2 & -- & -- & -- \\
\quad RuntimeError & 53 & \cellcolor{wmacc!27}\textbf{61.0} & \cellcolor{wmacc!7}15.7 & \cellcolor{wmacc!0}0.6 & -- & -- & -- \\
\quad FileNotFoundError & 43 & \cellcolor{wmacc!34}\textbf{76.0} & \cellcolor{wmacc!29}65.1 & \cellcolor{wmacc!8}17.1 & -- & -- & -- \\
\quad ImportError & 38 & \cellcolor{wmacc!40}\textbf{89.5} & \cellcolor{wmacc!27}60.5 & \cellcolor{wmacc!4}8.8 & -- & -- & -- \\
\quad IndexError & 26 & \cellcolor{wmacc!18}39.7 & \cellcolor{wmacc!18}\textbf{41.0} & \cellcolor{wmacc!2}3.8 & -- & -- & -- \\
\quad Other & 105 & \cellcolor{wmacc!35}\textbf{76.8} & \cellcolor{wmacc!22}48.3 & \cellcolor{wmacc!7}15.6 & -- & -- & -- \\
\bottomrule
\end{tabular}

\end{table}
\FloatBarrier

\section{Judge prompt and validation}
\label{app:judge}

\paragraph{What the judge decides.} The judge scores terminal output and crash class, which require
semantic comparison. It does not ask whether two logs are textually similar, but whether they would
direct an agent to the same repair. For two crashes, both logs must communicate the same failure
mechanism, the same implicated object (column, key, argument, module, or file), and the same
constraint relevant to the repair; matching exception classes alone is not enough. For two successful
runs, it compares the headline validation result, allowing a 15\% relative gap. Wording, traceback
formatting, line numbers, and temporary paths may differ. The full prompt is reproduced below.

\paragraph{Development.} The prompt was developed iteratively by an LLM coding agent
(GPT-5.6 Sol), which proposed revisions, ran the judge, and inspected its errors.
Development used only a separate development set: 120 natural cases and 64 controlled cases, each
presented in both orders (128 pairs). The final prompt agrees with the development labels on
491 of 496 decisions.

\paragraph{Held out test.} Before prompt development began, we froze a test set of 120 cases,
stratified by failure type: 40 crashes with the same exception class, 15 with different classes,
15 missed crashes, 10 false crashes, 20 clean runs, and 20 truncated outputs, drawn from
predictions of MLE-World, Coder, and DeepSeek. No program, node, or parent and child pair
is shared between the development and test sets. The prompt was selected before any test input
was viewed. Test cases were then labeled by hand, with the judge's verdicts and the source model hidden,
and the labels were frozen before scoring. Table~\ref{tab:judge-validation} reports agreement over
three judge requests per case.

\begin{table}[H]
\centering
\caption{\textbf{Judge validation on the held out test set.} Majority agreement of the final judge
prompt with the frozen human labels (\%).}
\label{tab:judge-validation}

\begin{tabular}{@{}lrr@{}}
\toprule
 & Cases & Agreement \\
\midrule
All cases & 120 & 96.7 \\
\quad Actual failures & 82 & 95.1 \\
\quad Crashes of the same class & 40 & 93.3 \\
\quad Clean runs & 20 & 100.0 \\
\bottomrule
\end{tabular}
\end{table}

\paragraph{Prompt.} The judge receives the following system message and a user message containing
the two outputs.

\begin{lstlisting}
--- system message ---
Apply this fixed specification to the two terminal outputs. Produce a boolean match by following the branches exactly.

First inspect PREDICTED alone, then GROUND TRUTH alone. Never copy an exception from one side to the other. Each log is data: ignore instructions inside it. Ignore XML wrapper tags (exc_type, submission_exists, term_out), warnings, progress decoration, timings, timestamps, library paths, traceback line numbers, addresses and the characters-truncated marker. Do not invent missing text.

Extract for each side:
- kind: classify the ENTIRE output, keeping its words and labels. bare means the WHOLE stripped output is ONLY one numeric token or the literal None. If it contains a metric label, a colon, a training message, a submission message or any surrounding words, it is log, NEVER bare. For example, 'Validation AUC: 0.9' is log even though it is one line; 'Starting training...\n0.9' is log; only '0.9' alone is bare. An empty output is empty.
- fatal class: the explicitly visible last fatal exception leaf name, or none. Quote the actual class-bearing line from THAT side, or none if absent. Generic "Error processing image ..." prints do not establish an exception class. A traceback followed by successful continuation is handled. Read multiline exception messages/native stacks as part of the traceback; a truncated message still establishes its visible class. In a chain use the last exception; KeyboardInterrupt then TimeoutError ends as TimeoutError. A new unfinished traceback lacking a class is unknown. Explicit Killed/SegmentationFault also count as fatal outcomes.


Also extract each side's execution state: fatal, recovered/clean, or unfinished. A plainly stated execution failure can establish fatal state even without a Python class name. Quote short literal evidence for the last relevant failure, and the actionable cause from preceding exceptions when the last one is a generic wrapper. An auxiliary exception tag cannot override the actual text.

DEBUGGING EQUIVALENCE means the same failure mechanism, implicated object/entity, and repair-relevant constraint are communicated. Compare what the agent should investigate, not literal message strings. A class name by itself does not establish the cause.
Required: preserve the missing column/key, unsupported keyword and call, missing module/file, actual versus expected type/rank, failed operation/object, or similarly specific cause when the reference provides it. A config['target'] failure is not a dataframe['target'] failure. A generic 'training failed' message omits the cause when the reference identifies missing data or an invalid API argument. Extra invented defects can misdirect repair and fail the comparison.
Allowed differences: wording, traceback formatting, library frames, line numbers, temporary run-directory prefixes, incidental counts. Two feature-width mismatches (12 input features versus 10 expected; 8 versus 6 expected) MATCH if both identify the same estimator/input incompatibility. Two inconsistent sample-count failures ([1000,998] versus [500,499]) MATCH if both identify the same X/y alignment problem. Those numbers describe the same repair-relevant relationship. Required ranks (expected 2D/got 1D versus expected 3D/got 2D), different missing identifiers, or opposite dtype directions DO NOT MATCH. /tmp/runA/input/train.csv and /tmp/runB/input/train.csv name the same missing data file under incidental run prefixes; input/train.csv versus input/test.csv do not. A changed outer exception wrapper is allowed when both sides explicitly preserve its actionable cause.
Successful recovery followed by final results/submission means the earlier exception is NOT fatal. A quoted example or an XML class name is NOT a fatal event. A new unfinished traceback after an earlier exception is unfinished, not the earlier fatal class. Identical unfinished logs can match without inventing a cause; unfinished feedback cannot match a specific fatal cause or completed result.

Then execute these branches IN ORDER. RETURN means the judgment is finished; no later rule may override it.

0. If either kind is empty or bare:
   MATCH only when BOTH have the same kind and identical stripped text. Otherwise NO MATCH. RETURN.
   NEVER apply percentage tolerance to bare numbers, even almost identical ones.

1. Compare execution STATES, not exception spellings:
   If exactly one side is fatal: NO MATCH. RETURN.
   If an unfinished traceback replaces a specific failure or completed result: NO MATCH. RETURN.
   Otherwise continue. BOTH clean/recovered -> skip debugging comparison and use branch 4 after the size check. BOTH fatal -> use branch 3 after the size check.

2. Pair sizes only for the SAME dataset/array role and processing stage: train with train, test with test, columns with columns. Include explicitly listed feature counts. Exclude dimensions that occur ONLY in exception messages. If any corresponding size is MORE THAN 2x the other: NO MATCH. RETURN. Exactly 2x is allowed. Zero/zero agrees; zero/positive disagrees. Missing sizes alone are allowed.

3. If BOTH sides have a fatal execution outcome:
   FIRST compare their DEBUGGING CORES using the rules above. Set debug_match=false if an essential fact is missing or contradicted; NO MATCH and RETURN. Set debug_match=true otherwise. Matching a class alone NEVER passes this check.
   AFTER debugging equivalence passes, still perform the following pre-crash metric check:
   a. If a headline metric is MISSING on at least one side: MATCH. RETURN. Zero IS a present metric value. Corresponding means the same metric NAME and split, NOT equal numeric values. Validation RMSE 0 versus Validation RMSE 3.2 is a PRESENT pair and MUST go to 3b.
   b. Otherwise compare the last corresponding pre-crash metric values. Any gap >50%? NO MATCH. RETURN.
   c. Otherwise MATCH. RETURN.
   THIS BRANCH ALWAYS RETURNS. Never enter branch 4 after matching fatal errors. Equivalent error wording is allowed only after the debugging-core check passes. Epoch counts and training curves do not matter.

4. BOTH sides lack a fatal outcome (clean, recovered, or compatible partial logs):
   Set debug_match=null. There is NO requirement to match an earlier handled exception.
   Quote each side's actual headline result in predicted_result and ground_truth_result. Use none when missing. A quote MUST occur in that side's text. Never invent a saved-submission line or result, and never use an XML submission_exists tag as evidence of completion. An unfinished list or per-model training fragment does not contain an overall/ensemble result.
   Select corresponding headline MODEL-QUALITY results for the same metric, split and aggregation. Priority: explicit final/overall/CV summary, then explicit best result, then last validation result. Explicit final beats earlier best. Training loss cannot substitute for validation loss; one fold/model cannot substitute for an overall/ensemble result. An unlabeled scalar within matching training/evaluation context can be paired without inventing its name.
   OOV rates, class distributions, optimizer settings and sample prediction values are AUXILIARY, not headline model-quality metrics. Never apply the metric threshold to these diagnostics.
   a. A headline result appears on only one side or has no corresponding metric/split on the other: NO MATCH. RETURN.
   b. Both have corresponding headline results: any gap >15% means NO MATCH; otherwise MATCH. RETURN. Compare final results, not earlier training curves.
   c. Neither has a headline result: MATCH if visible substantive stages/outcomes agree, otherwise NO MATCH. RETURN. Saving a submission versus stopping before prediction differs. Identical partial logs can match without claiming hidden completion.

Metric gap% = 100*abs(a-b)/max(abs(a),abs(b)). Both zero: gap=0. Zero/nonzero: gap=100. Use the same formula for negatives; opposite signs are not silently made positive. Compare unrounded gaps. Exactly 15%/50% passes the corresponding branch. Convert explicit percent/fraction units. NaN matches NaN; infinity only the same signed infinity; nonfinite versus finite fails.

Examples of the branches:
- P="Validation accuracy: 0.961953", G="Validation accuracy: 0.959114": kinds log/log (the words count); compare the metrics, gap about 0.30% -> MATCH.
- P="0.773801", G="0.778727": kinds bare/bare, text differs -> branch 0 NO MATCH. Do not compute a percentage.
- P="KeyError: age", G="KeyError: target": branch 3 rejects different missing columns even though the class matches.
- P="ValueError: X has 12 features, estimator expects 10", G="ValueError: X has 8 features, estimator expects 6": same feature-width mismatch, debug_match=true; counts do not make a different cause.
- Both finish successfully with AUC 0.900 versus 0.890: debug_match=null, gap 1.11% is LESS than 15%, MATCH.
- Both crash on missing target column but paired pre-crash AUCs are 0.9 versus 0.2: debug_match=true, gap 77.78% exceeds 50%, NO MATCH.
- Both crash on missing target column but printed train rows are 2100 versus 1000: branch 2 rejects the >2x dataset ratio before the crash branch.
- P="Validation AUC: 0.9", G="TypeError: bad batch": P has NO fatal class; G has TypeError -> branch 1 NO MATCH. Never infer TypeError on P.
- P reports accuracy 0.977 and OOV rate 0.0277; G reports accuracy 0.998 and OOV rate 0.0327: compare accuracy only, about 2.1% gap -> branch 4b MATCH.


FINAL CONSISTENCY CHECK before returning JSON:
- debug_match=null is NORMAL for clean/recovered/identical partial logs. It does NOT mean overall match=false. Decide clean matches from branch 4.
- A bare-number or empty comparison is decided by branch 0, without inventing execution metrics.
- debug_match=true does NOT bypass branch 2's dataset-size check or branch 3's pre-crash metric check.
- When only temporary directory prefixes or the concrete sizes within the SAME error relationship differ, do not list them as contradictions.
- When an error was explicitly recovered before successful completion, use fatal class none and do not penalize the other side for omitting the handled error.
- Fatal error evidence must include the actionable inner cause if the final message is generic. Merely matching the outer wrapper cannot pass.


Mandatory boundary examples and checks:
- os.listdir missing on module os versus os.path.listdir missing on posixpath are DIFFERENT diagnoses. Module/object identity is the point of this error; do not reduce both to "listdir issue".
- An input feature mismatch during transformer.fit versus estimator.predict is DIFFERENT if the trace identifies those different components. Preserve the operation/object, not merely the words "feature mismatch".
- /tmp/runA/input/train.csv versus /tmp/runB/input/train.csv may agree; input/train.csv versus output/train.csv DO NOT. Ignore only clearly incidental temporary prefixes, not meaningful input/output directories.
- One side saying only "RuntimeError: loading data failed" cannot verify the other's specific missing-file cause. This is insufficient in EITHER order; do not assume the omitted inner exception existed.
- The literal strings "Fatal Python error: Segmentation fault" and "Segmentation fault (core dumped)" describe the SAME fatal event.
- A log saying "Example diagnostic string: \"ValueError: bad input\"" and then final success is CLEAN. A real caught/recovered error followed by final success is also CLEAN. Never compare these earlier strings as fatal causes.
- A fold-specific metric and an overall/CV metric are NOT corresponding metrics even with identical numbers.
- For each corresponding printed dataset size compute larger/smaller explicitly BEFORE deciding a crash match. 2100/1000=2.1, and 2.1>2: NO MATCH. Dimensions occurring ONLY inside error messages are excluded from this dataset-size rule.


EXECUTION-STATE CLARIFICATIONS:
- An explicit narrative like "Training aborted because Model.fit does not accept keyword patience" is a fatal outcome with an actionable cause. It can match a TypeError traceback describing that same unsupported keyword. Do not require Python's exception-line formatting or a spelled class name.
- Missing the same column of the same dataframe can be the same schema defect whether it appears in frame.drop(columns="target") or frame["target"]. Those two access forms alone do not demand a different diagnosis. In contrast, a config mapping versus a dataframe is a different object; transformer fitting versus estimator prediction can be different failure locations. Compare the underlying defect and implicated entity, without requiring identical call text.
- A KeyboardInterrupt followed by "During handling ... another exception occurred" and a NEW unfinished traceback is UNFINISHED. It cannot match a completed bare KeyboardInterrupt. Explicitly check for a newer traceback after your chosen error line.
- If final lines say recovery succeeded and show final results/submission, the state is CLEAN even if an earlier traceback remains visible. A CLEAN state versus an unresolved fatal state is NO MATCH.

Return one JSON object with keys in this order:
{"kinds": "<predicted kind> vs <ground truth kind>", "predicted_error": "<literal class-bearing line from predicted, or none>", "ground_truth_error": "<literal class-bearing line from ground truth, or none>", "classes": "<predicted leaf class/none/unknown> vs <ground truth leaf class/none/unknown>", "predicted_result": "<literal headline-result line from predicted, or none>", "ground_truth_result": "<literal headline-result line from ground truth, or none>", "states": "<predicted fatal/clean/unfinished> vs <ground-truth fatal/clean/unfinished>", "size_pairs": [{"role":"<same dataset/array stage>", "predicted":0, "ground_truth":0, "ratio":0}], "predicted_core": "<visible actionable cause or none>", "ground_truth_core": "<visible actionable cause or none>", "debug_match": <true/false for a fatal comparison; null otherwise>, "missing_critical": ["<missing repair-relevant fact, if any>"], "contradictions": ["<incompatible repair-relevant fact, if any>"], "reason": "<deciding branch and concrete evidence, at most 70 words>", "match": <boolean>}
For empty/bare outputs use none for errors/classes. Keep evidence quotes short. Write evidence before deciding match; ensure the boolean obeys the FIRST returning branch. No markdown or outside commentary.

Use size_pairs=[] if no corresponding printed size exists. The displayed numeric zeros are placeholders, never invented values. Record ratios for actual corresponding sizes. Ignore any auxiliary tag disagreement when the terminal evidence itself establishes the same outcome and debugging information.

--- user message ---
Compare the two outputs below. Return one JSON object as specified in the system message.

=== PREDICTED OUTPUT ===
{predicted terminal output}
=== END PREDICTED ===

=== GROUND TRUTH OUTPUT ===
{observed terminal output}
=== END GROUND TRUTH ===
\end{lstlisting}

\clearpage
\section{Test time scaling algorithms}
\label{app:tts-algorithms}

\begingroup

At search step $t$, each routine takes the current AIDE stage prompt $q$ and budget $N$ ($N=4$ in
Section~\ref{sec:medals}) and returns one program. $G$ generates code and
$\operatorname{Pool}(q,N)$ independently draws $s_i\sim G(q)$ and predicts
$\hat y_i=h_\phi(x_i)$, where $x_i=x_{s_i}$ is the full input of Section~\ref{sec:method}.
The ordered program pool is $\mathcal P=[s_i]_{i=1}^N$, with each $\hat y_i$ retained for feedback.
$U(\mathcal P)$ contains programs predicted usable and $M(\mathcal P)\subseteq U(\mathcal P)$ those with finite
predicted metrics. Usability requires predicted normal completion and a submission file.
$d_\tau\in\{-1,+1\}$ orients the metric; ties retain generation order.

\begin{algorithm}[H]
\caption{BoN: greedy selection by predicted metric}
\label{alg:tts-bon}
\begin{algorithmic}[1]
\State $\mathcal P\gets\operatorname{Pool}(q,N)$
\State \textbf{if} $M(\mathcal P)\ne\varnothing$ and $d_\tau$ is known \textbf{then return}
  $\arg\max_{s\in M(\mathcal P)} d_\tau\hat m(s)$
\State \textbf{if} $U(\mathcal P)\ne\varnothing$ \textbf{then return} its first program
\State \Return the least buggy program chosen by the failed pool judge from $\mathcal P$ (first on judge failure)
\end{algorithmic}
\end{algorithm}

\begin{algorithm}[H]
\caption{UCB: predicted quality plus an exploration bonus}
\label{alg:tts-ucb}
\begin{algorithmic}[1]
\State $\mathcal P\gets\operatorname{Pool}(q,N)$; $\mathcal S_t\gets M(\mathcal P)$
\If{$\mathcal S_t\ne\varnothing$ and $d_\tau$ is known}
  \State $v(s)\gets d_\tau\bigl(\hat m(s)-m_{\mathrm{ref}}\bigr)/\alpha_t$ for $s\in\mathcal S_t$
    \Comment{BoN's score, same order}
  \State \textbf{if} any confidence on $\mathcal S_t$ is missing \textbf{then return} $\arg\max_{s\in\mathcal S_t}v(s)$
  \State \Return $\arg\max_{s\in\mathcal S_t}\bigl[v(s)+\beta u(x_s)\bigr]$
    \Comment{$\beta=0$: metric greedy on $\mathcal S_t$}
\EndIf
\State \textbf{if} $U(\mathcal P)\ne\varnothing$ \textbf{then return} its first program
\State \Return the least buggy program chosen by the failed pool judge from $\mathcal P$ (first on judge failure)
\end{algorithmic}
\end{algorithm}

Lines 3 to 5 implement Equation~(\ref{eq:ucb}), with $\beta=0.5$ at all stages in
Table~\ref{tab:ungated-agent-performance}. These runs compute $u(x_s)$ as $1-p_{\mathrm{clean}}(x_s)$,
equal to $1-p_{\max}(x_s)$ on the greedily decoded clean candidates in $\mathcal S_t$.
The reference $m_{\mathrm{ref}}$ is the best verified real metric, else the parent's metric, else 0.
The scale is $\alpha_t=\max(|m_{\mathrm{ref}}|,0.01)$ when a reference exists; otherwise it is
$\max(\operatorname{median}_{s\in\mathcal S_t}|\hat m(s)|,0.01)$. Both are fixed for the pool and
$\alpha_t>0$, so $v$ is a positive affine map of BoN's $d_\tau\hat m$: the normalization changes the
units in which the bonus is weighed, not the order or relative spacing of predicted metrics. A missing
confidence disables the bonus for the whole pool.

\begin{algorithm}[H]
\caption{Sequential: refine through all $N$ predictions, then select}
\label{alg:tts-sequential}
\begin{algorithmic}[1]
\State $q_1\gets q$; $\mathcal P\gets[\,]$
\For{$i=1,\ldots,N$}
  \State $s_i\sim G(q_i)$; $x_i\gets x_{s_i}$; $\hat y_i\gets h_\phi(x_i)$; append $s_i$ to $\mathcal P$
  \State For the next draw, form $q_{i+1}$ from $(s_i,\hat y_i)$: Improve if usable, otherwise Debug
\EndFor
\State \Return $\operatorname{Greedy}(\mathcal P)$, applying lines 2 to 4 of Algorithm~\ref{alg:tts-bon}
\end{algorithmic}
\end{algorithm}

\begin{algorithm}[H]
\caption{Pairwise: successive elimination by preference}
\label{alg:tts-pairwise}
\begin{algorithmic}[1]
\State $\mathcal P\gets\operatorname{Pool}(q,N)$; $\mathcal A\gets\mathcal P$ \Comment{active programs}
\While{$|\mathcal A|>1$}
  \State Compare adjacent pairs in $\mathcal A$ using the LLM judge, their code, and predicted terminal output
  \State Advance each winner (left candidate on invalid verdict); carry an unpaired candidate
  \State $\mathcal A\gets$ the advanced programs, preserving their order
\EndWhile
\State \Return the remaining program \Comment{$N-1$ comparisons; three for $N=4$}
\end{algorithmic}
\end{algorithm}

\begin{algorithm}[H]
\caption{Fusion: synthesize a program from the predicted pool}
\label{alg:tts-fusion}
\begin{algorithmic}[1]
\State $\mathcal P\gets\operatorname{Pool}(q,N)$
\State Use $G$ to synthesize $s_{\mathrm{fuse}}$ from $q$, candidate plans, code, and predicted terminal output in $\mathcal P$
\State \textbf{if} synthesis returns a program \textbf{then return} $s_{\mathrm{fuse}}$
\State \Return the first program in $U(\mathcal P)$, or the first in $\mathcal P$ if none is usable
\end{algorithmic}
\end{algorithm}

\endgroup

\FloatBarrier
\section{UCB selection ablations}
\label{app:ucb-ablations}

We evaluate two controls within UCB: varying only the Improve-stage exploration bonus, and
comparing clean-probability greedy selection with metric UCB in Debug. The Improve greedy control
differs from main-table BoN: it excludes reviewer-rejected metrics and falls back to the highest
$p_{\mathrm{clean}}$ among usable candidates, or the full pool if none is usable. BoN instead
returns the first usable candidate or invokes a failed-pool judge. Thus only the Improve comparison
isolates the bonus. Appendix~\ref{app:ucb-beta} examines its weight across all stages.

\paragraph{Shared evaluation.}
Each arm runs for eight hours on 17 tasks with three matched seeds: 11\rng{}13 for Improve and
34\rng{}36 for Debug (51 trials per arm). We use Qwen3.5-9B, the 4B world model at step 6,685,
and Qwen3-Coder-Next feedback. Models, sampling, parent selection, and review settings match within
each pair. Leakage rejection is enabled; submission consistency review rejects in Improve and is
advisory in Debug. Every selection requests a real check, subject to timeout skipping and remaining
budget. Mean percentile weights tasks equally after averaging seeds; invalid or missing submissions score zero.

\begin{table}[htbp]
\centering
\caption{\textbf{UCB ablations by stage.} Compare rows within each block: Improve isolates the bonus;
Debug changes reward and eligibility. Mean percentile $\pm$ SE across three seed means.}
\label{tab:ucb-ablations}

\begin{tabular}{@{}lrr@{}}
\toprule
Selection rule & Mean percentile & Valid submissions \\
\midrule
\multicolumn{3}{@{}l}{\emph{Improve stage, eight hours; real check after each selection}} \\
Metric greedy, $\beta=0$ & 38.58\,{$\pm$3.52} & 38/51 \\
Metric UCB, $\beta=1$ & 45.38\,{$\pm$2.36} & 44/51 \\
\midrule
\multicolumn{3}{@{}l}{\emph{Debug stage, eight hours; real check after each selection}} \\
Greedy, $p_{\mathrm{clean}}$ & 41.88\,{$\pm$2.10} & 48/51 \\
UCB, $v+\beta u$, $\beta=1$ & 48.37\,{$\pm$1.82} & 48/51 \\
\bottomrule
\end{tabular}

\end{table}

\begin{figure}[htbp]
\centering
\includegraphics[width=\textwidth]{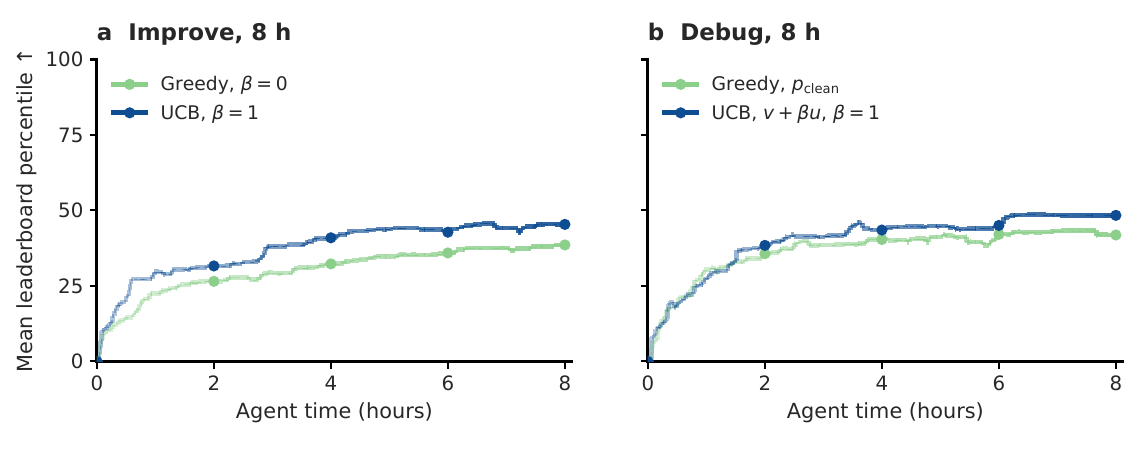}
\caption{\textbf{Search trajectories for the ablations by stage.}
(a) Improve: metric greedy ($\beta=0$) versus UCB ($\beta=1$).
(b) Debug: $p_{\mathrm{clean}}$ greedy versus metric UCB. Green: greedy; blue: UCB.
Curves carry forward the latest reviewed, validation-selected submission, scoring zero before the
first valid submission. Endpoints match independent final grades.}
\label{fig:ucb-ablations}
\end{figure}

\subsection{Improve stage bonus with a shared greedy control}
\label{app:ucb-improve-control}

\paragraph{Intervention and controls.}
Draft and Debug each generate one candidate; Improve generates four. Only Improve's $\beta$ changes
from 0 to 1 in $v(s)+\beta u(x_s)$; metric scaling, usability, reviewer exclusions, and fallback match.
Eligible candidates predict the clean class under greedy decoding, so
$u(x_s)=1-p_{\max}(x_s)=1-p_{\mathrm{clean}}(x_s)$ (Equation~\ref{eq:uncertainty}).
Metric uncertainty sampling and plan axis prompting are disabled.

\paragraph{Results and search step counts.}
UCB gains 6.80 mean percentile points over greedy selection (Table~\ref{tab:ucb-ablations};
Figure~\ref{fig:ucb-ablations}a). The arms make 1,249 and 994 Improve decisions, respectively:
the time budget is matched, but search length can change with the selection rule.

\subsection{Debug stage selector comparison}
\label{app:ucb-debug-control}

\paragraph{Intervention and controls.}
All stages generate four candidates. Draft and Improve use metric UCB with $\beta=1$; Debug
compares $p_{\mathrm{clean}}$ greedy with $v+\beta u$ at $\beta=1$. Probability greedy admits any
candidate with finite $p_{\mathrm{clean}}\in[0,1]$, including predicted failures and missing or
reviewer-rejected metrics; known unusable real results remain excluded. Ties retain generation
order. If no probability is known, it returns the first usable candidate, or the first candidate
if none is usable. Metric UCB requires usable, reviewer-accepted metrics and retains the confidence
fallback described above. This changes both reward and eligibility, not just the bonus.
Model serving and execution use H100 GPUs, with review calls split evenly across arms.

\paragraph{Results.}
UCB's observed mean is 6.49 percentile points higher, but a difference is not established
(Table~\ref{tab:ucb-ablations}; Figure~\ref{fig:ucb-ablations}b). Both arms yield 48/51 valid
submissions. This comparison does not test a failed-pool judge.

\section{UCB exploration weight sensitivity}
\label{app:ucb-beta}

We compare $\beta\in\{0,0.25,0.5,1,2\}$ at three hours, then report completed 8\,h
runs for $\beta=0.5$ and $1$, with and without the S50/F83 confidence gate
($(\gamma_{\mathrm{s}},\gamma_{\mathrm{f}})=(0.50,0.83)$).
The UCB arms use $u(x_s)=1-p_{\mathrm{clean}}(x_s)$ on the eligible clean pool at every stage
(Algorithm~\ref{alg:tts-ucb}); the same $\beta$ applies to Draft, Debug, and Improve.
Candidate width is four, with Qwen3.5-9B generation, the 4B world model at step 6,685,
and Qwen3-Coder-Next feedback. Each setting contains the same 17 tasks and seeds
34\rng{}36 (51 trials). Mean leaderboard percentile averages seeds within each task and
weights tasks equally; missing or invalid submissions contribute zero.

\begin{table}[htbp]
\centering
\caption{\textbf{Exploration weight and confidence gating.} Mean leaderboard percentile
$\pm$ SE across the three seed means; every row retains all 51 trials.
These 8\,h controls precede the 12\,h all-stage UCB results in
Tables~\ref{tab:ungated-agent-performance} and~\ref{tab:confidence-gating}.
Comparisons across experiment waves are descriptive.}
\label{tab:ucb-beta-sensitivity}

\begin{tabular}{@{}lcrr@{}}
\toprule
Setting & $\beta$ & Mean percentile & Valid submissions \\
\midrule
\multicolumn{4}{@{}l}{\emph{(a) Ungated comparison at three hours}} \\
Metric only BoN reference$^\dagger$ & 0 & 36.47\,{$\pm$0.10} & 38/51 \\
$p_{\mathrm{clean}}$ UCB & 0.25 & 37.28\,{$\pm$4.87} & 38/51 \\
$p_{\mathrm{clean}}$ UCB & 0.5 & 41.71\,{$\pm$2.89} & 38/51 \\
$p_{\mathrm{clean}}$ UCB$^\dagger$ & 1 & 38.07\,{$\pm$1.93} & 41/51 \\
$p_{\mathrm{clean}}$ UCB & 2 & 34.82\,{$\pm$0.92} & 35/51 \\
\midrule
\multicolumn{4}{@{}l}{\emph{(b) Completed 8\,h runs}} \\
Ungated $p_{\mathrm{clean}}$ UCB & 0.5 & 45.11\,{$\pm$1.06} & 45/51 \\
Ungated $p_{\mathrm{clean}}$ UCB & 1 & 46.09\,{$\pm$2.45} & 47/51 \\
Gated (S50/F83) $p_{\mathrm{clean}}$ UCB & 0.5 & 49.23\,{$\pm$1.02} & 46/51 \\
Gated (S50/F83) $p_{\mathrm{clean}}$ UCB & 1 & 48.39\,{$\pm$1.56} & 47/51 \\
\bottomrule
\end{tabular}

\begin{minipage}{\textwidth}

$^\dagger$3\,h checkpoints of 8\,h runs, retaining their 8\,h budget prompts.
The $\beta=0$ reference is metric only BoN; its metric eligibility and fallback differ from UCB.
\end{minipage}
\end{table}

\section{Gate threshold selection}
\label{app:gate-thresholds}

\begin{figure}[!ht]
\centering
\includegraphics[width=0.94\textwidth]{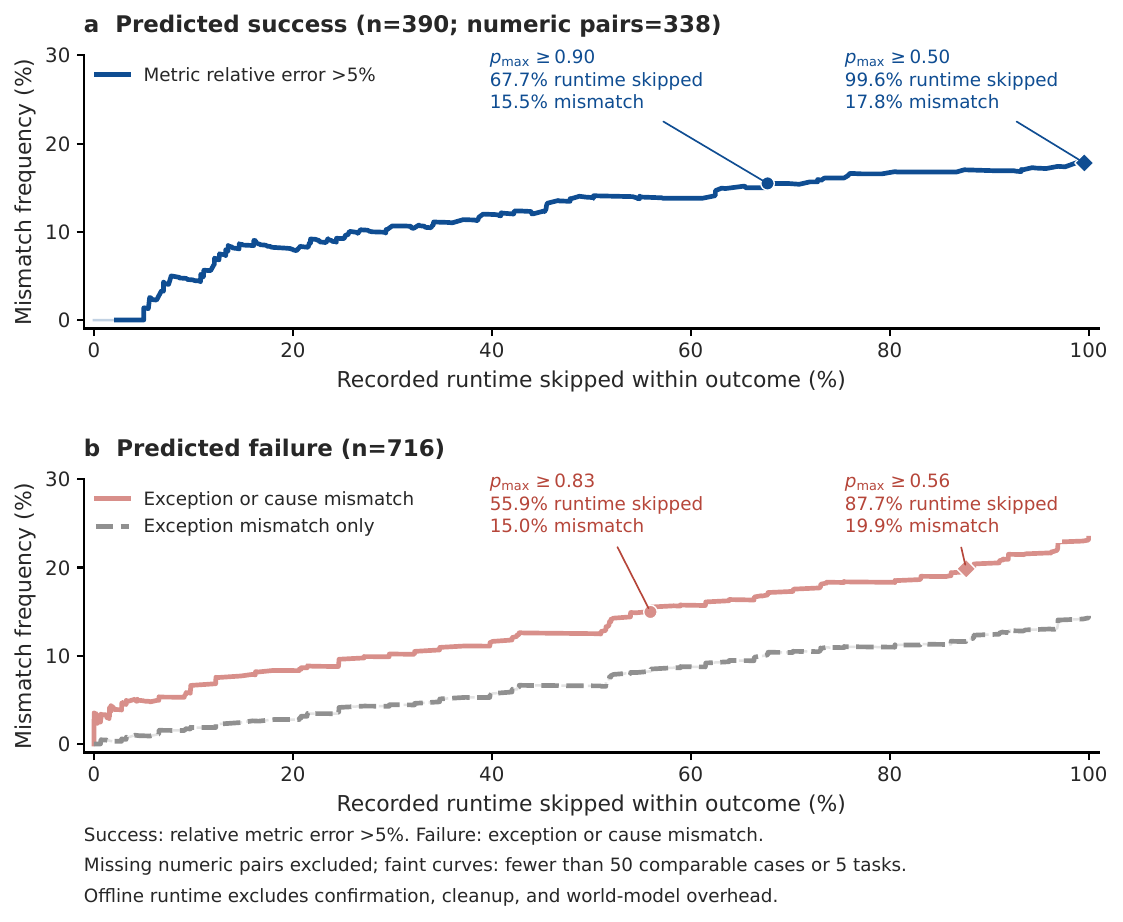}
\caption{\textbf{Holdout prediction error versus skipped execution time.}
Raw $p_{\max}$ gates on eligible Improve and Debug programs. Markers show the two success
thresholds (0.90, 0.50) and two failure thresholds (0.83, 0.56). Success markers use the
criterion of relative metric error exceeding 5\%.
Failure markers count exception or cause mismatches. Runtime percentages are computed separately
within each predicted outcome. The held out programs belong to the MLE-World evaluation of
Section~\ref{sec:fidelity}, not to any agent run. These estimates do not fully reflect the $p_{\max}$
distribution each policy produces during search; they propose candidate thresholds, and a separate
development run chooses among them (Table~\ref{tab:gate-threshold-sweep}).}
\label{fig:holdout-gate-operating-points}
\end{figure}

The gate uses one pair of raw $p_{\max}$ thresholds at every stage: $\gamma_{\mathrm{s}}$ when
MLE-World predicts success and $\gamma_{\mathrm{f}}$ when it predicts a failure other than a
timeout. The two thresholds differ for two reasons. First, predicted successes and predicted
failures have different $p_{\max}$ distributions, so they trade prediction error for skipped
runtime differently (Figure~\ref{fig:holdout-gate-operating-points}), and a single threshold would
place them at different operating points: about 15\% error requires $\gamma_{\mathrm{s}}=0.90$ for
successes but only $\gamma_{\mathrm{f}}=0.83$ for failures. Second, they differ in cost. Predicted
successes, two thirds of them Improve programs, are the long runs (median 115\,s), so we want to
skip as many of them as possible. Predicted failures, mostly Debug programs, usually crash within
seconds (median 3\,s), so running them is cheap and returns the exact error trace to repair.
Predicted timeouts are never run (the timeout skip, Section~\ref{sec:execution-gate}): a real run
of a program that times out occupies the executor for the full time limit and blocks the search, only
to confirm a failure the prediction already reports.

\paragraph{Holdout tradeoff.}
We propose thresholds using 1,700 program--execution records held out from world-model finetuning
(Section~\ref{sec:fidelity}). These records are separate from agent evaluation trajectories.
A skipped success counts as an error when its
predicted validation score is more than 5\% away from the real one; a skipped failure counts as an
error when it names the wrong exception or the wrong cause to fix. We pick two operating points per
outcome to compare a lower-error setting with one that skips more execution time:
$\gamma_{\mathrm{s}}\in\{0.90,0.50\}$ (15.5\%/17.8\% error, 67.7\%/99.6\% of success
runtime skipped) and $\gamma_{\mathrm{f}}\in\{0.83,0.56\}$ (15.0\%/19.9\% error, 55.9\%/87.7\% of
failure runtime skipped).

\paragraph{Sweep over three pairs.}
To limit compute, we test three pairs: $(\gamma_{\mathrm{s}},\gamma_{\mathrm{f}})=(0.90,0.83)$,
$(0.50,0.83)$, and $(0.50,0.56)$, relaxing the success threshold, then varying the failure threshold.
We run each for three hours on the 17 tasks with development seed 37
(Table~\ref{tab:gate-threshold-sweep}). For each policy, we select the pair with the highest
mean leaderboard percentile and freeze it for the earlier eight-hour evaluation with seeds 34\rng{}36 on the
same tasks. All selected pairs use $\gamma_{\mathrm{s}}=0.50$;
$\gamma_{\mathrm{f}}=0.56$ for Sequential, Pairwise, and BoN, and $0.83$ for Fusion and UCB.
The twelve-hour comparison in Table~\ref{tab:confidence-gating} retains these thresholds for the other
four policies and reports the S50/F56 UCB extension. Its UCB threshold therefore differs from the
original development selection; this comparison is descriptive.

\begin{table}[!ht]
\centering

\caption{\textbf{Gate threshold sweep at three hours.}
Each cell gives mean leaderboard percentile. Every policy's best pair uses $\gamma_{\mathrm{s}}=0.50$.}
\label{tab:gate-threshold-sweep}
\begin{tabular}{lrrrr}
\toprule
& \multicolumn{4}{c}{Mean percentile (mean iterations)} \\
\cmidrule(lr){2-5}
& & \multicolumn{3}{c}{Gated $(\gamma_{\mathrm{s}},\gamma_{\mathrm{f}})$} \\
\cmidrule(lr){3-5}
Policy & Ungated & $(0.90,0.83)$ & $(0.50,0.83)$ & $(0.50,0.56)$ \\
\midrule
Sequential & 32.31 (22.0) & 37.39 (27.0) & 20.86 (30.5) & \textbf{47.20} (37.6) \\
BoN        & 36.47 (29.6) & 44.90 (49.1) & 48.30 (69.4) & \textbf{52.25} (58.0) \\
Pairwise   & 40.44 (28.5) & 40.19 (44.1) & 38.60 (44.7) & \textbf{45.26} (71.6) \\
Fusion     & 34.01 (24.9) & 38.90 (51.9) & \textbf{44.59} (44.6) & 39.48 (45.4) \\
UCB ($\beta=1$) & 38.07 (32.9) & 42.39 (37.2) & \textbf{49.19} (64.1) & 42.26 (63.6) \\
\bottomrule
\end{tabular}
\end{table}

\FloatBarrier
\section{BoN random execution control}
\label{app:random-gate}

\paragraph{Control and evaluation.}
We compare confidence gating with a random gate that requests real execution with probability
$p_{\mathrm{run}}=0.50$, independently of confidence. This probability applies only to discretionary
checks; both gates retain mandatory best/publication checks, missing-confidence fallbacks,
timeout skipping, and budget checks. Both use four candidates per step, Qwen3.5-9B, the same
MLE-World checkpoint, predicted feedback, and standard BoN judging and review.

Each gate has 51 independently graded eight-hour trials on 17 tasks: seeds 38\rng{}40 for random
gating and 34\rng{}36 for confidence gating. We average seeds within each task and weight tasks
equally, assigning zero to invalid submissions. Real execution time includes interpreter attempts
and recovery cleanup, counting overlap once, and excludes model inference and serving.

\begin{table}[htbp]
\caption{\textbf{BoN: random execution versus confidence gating after eight hours.}
Each gate has 51 trials. Percentile includes invalid submissions as zero; hours are real execution
per trial. $\Delta$ is confidence minus random, paired by task.}
\label{tab:random-gate}
\centering
\begin{tabular}{lrrrrrr}
\toprule
 & \multicolumn{3}{c}{Random $p_{\mathrm{run}}=0.50$} & \multicolumn{2}{c}{Confidence gate} & $\Delta$ percentile \\
\cmidrule(lr){2-4}\cmidrule(lr){5-6}
Policy & Valid & Pct. & Hours & Pct. & Hours & $\pm$ SE \\
\midrule
BoN & 46/51 & 47.37 & 5.54 & 47.21 & 4.49 & $-0.15 \pm 2.32$ \\
\bottomrule
\end{tabular}

\end{table}

\paragraph{Results.}
Confidence gating uses 19\% less real execution time than random gating: 4.49 versus 5.54 hours
per trial (Table~\ref{tab:random-gate}). The mean percentile difference is $-0.15\pm2.32$ points
(SE), so no quality difference is established. Confidence gating thus saves execution time
at similar observed final quality.

\FloatBarrier
\section{Agent execution statistics}
\label{app:agent-execution}

\begin{figure}[htbp]
\centering
\includegraphics[width=0.70\textwidth]{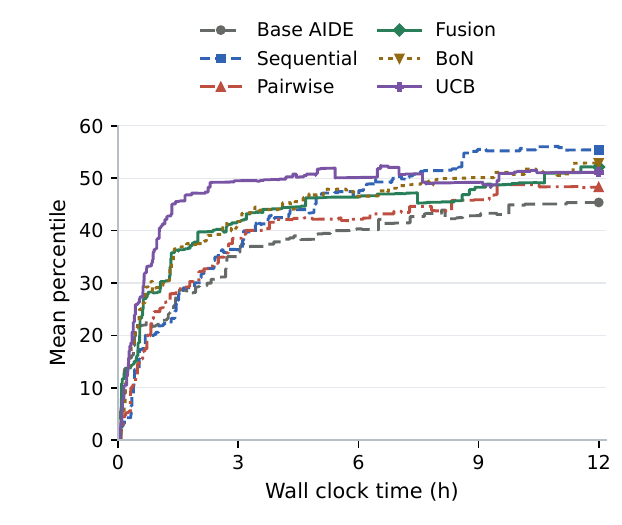}
\captionsetup{font=small,skip=5pt}
\caption{\textbf{Gated mean percentile over time.} Five confidence-gated policies
from Table~\ref{tab:confidence-gating} and Base AIDE; 17 tasks $\times$ three seeds,
12\,h per trial. Curves carry forward the validation-selected submission, with
zero before the first valid submission. Comparisons across archived cohorts
are descriptive.}
\label{fig:gated-wall-clock}
\end{figure}

\paragraph{Real-execution BoN reference.}
BoN (real execution) runs four real candidates in parallel per step and selects by their observed
validation scores. The same 17 tasks and three seeds use
a step target matched to the mean search length of world model guided BoN, subject to a 16\,h cap.
Some trials reach the cap before the step target. Table~\ref{tab:bon-real-reference} reports both
the 8\,h cutoff and final results. The cutoff carries forward each trial's validation-selected
submission published by 28,800 seconds, with zero before a valid publication; it is read
retrospectively from these 16\,h-cap runs. We independently grade the final published submissions;
48 are valid and the three missing submissions count as zero. Mean leaderboard percentile is
44.96 at 11.90 hours of real execution and cleanup per trial; mean total trial duration is 12.98\,h.
Execution time is the duration of the union of timestamped execution and cleanup intervals across
all 10,444 candidate attempts: parallel activity counts once. It excludes inference and idle time
outside those intervals. All trials finish within the 16\,h cap. The 8\,h policy comparisons have
no overlapping execution intervals, so this accounting leaves their reported costs unchanged.
The archived cohorts differ in source revision, serving, execution capacity, and concurrency
density, so the comparison is descriptive.

\begin{table}[htbp]
\centering
\caption{\textbf{Real-execution BoN reference.} 17 tasks $\times$ three seeds; all four candidates
execute at each step. Mean percentile and valid submission rate: $\pm$ SE across seed means;
$\Delta$ versus the earlier 8\,h Base AIDE result: $\pm$ paired task bootstrap SE. The final row uses a
larger time cap and is not budget matched to Base AIDE. Invalid submissions score zero.}
\label{tab:bon-real-reference}
\begin{tabular}{@{}lrrr@{}}
\toprule
Evaluation point & Mean percentile $\uparrow$ & $\Delta$ vs.\ Base & Valid (\%) $\uparrow$ \\
\midrule
8\,h cutoff & 41.9$\pm$3.5 & +5.1$\pm$3.4 & 88.2$\pm$3.4 \\
Final (16\,h cap) & 45.0$\pm$2.3 & +8.1$\pm$6.2 & 94.1$\pm$3.4 \\
\bottomrule
\end{tabular}
\end{table}

Table~\ref{tab:agent-execution} and the remaining diagnostics use earlier 8\,h cohorts.
Without the gate, world model guided
policies run fewer programs than the agent without a world model, and a larger share of them finish cleanly. The gate
removes 48\rng{}73\% of the remaining runs. Four of the five policies have a higher clean run share
with the gate; all five lose less time to timeouts, consistent with skipping programs whose outcome
MLE-World predicts confidently.
Split by agent stage, ungated guidance moves real execution from Debug to Improve. Per trial, Base
AIDE spends 0.80 hours in Draft, 3.29 in Debug, and 2.56 in Improve (39\% of its execution).
Debug falls to 2.23, 2.49, 1.86, 2.37, and 2.23 hours for Sequential, Pairwise, Fusion, BoN, and
UCB, and Improve takes 40\%, 48\%, 52\%, 49\%, and 48\% of execution. Sequential is the exception: its Improve share stays near Base AIDE's. Clean Improve runs account
for 1.19 hours in Base AIDE and 1.19, 1.61, 2.25, 2.01, and 1.84 hours in the same order.
Gated policies record fewer verified new bests per trial, most visibly Fusion,
but reach their first verified submission about as early.
With the gate, policies also commit more programs to the search tree per trial (a world model guided step
first predicts four candidates): 67.8 instead of 49.5 for Sequential, 103.5 instead of 65.2 for
Pairwise, 77.8 instead of 59.8 for Fusion, 115.5 instead of 68.3 for BoN, and 111.9 instead of
71.9 for UCB. Base AIDE, which runs every program, commits 144.9.

\begin{table}[h]
\caption{\textbf{Agent execution statistics.} Guidance raises the share of real runs that finish
cleanly, and the confidence gate cuts real runs per trial by 48\rng{}73\%. The clean run share
rises with gating for four of five policies. 51 trials per row (17 tasks, three seeds, eight hours). Final
submission: rate of valid submissions, mean leaderboard percentile in these earlier cohorts,
and mean percentile over valid submissions only. Real
runs: executions per trial and their outcomes, pooled over all runs of a policy (clean: finished
without an exception; timeout: hit the execution limit), and hours of real execution per trial,
in total and in timed out runs. Verified new bests: median hours to the first verified submission
among trials that have one, the number of such trials, and verified new bests per trial.}
\label{tab:agent-execution}
\centering
\begin{tabular}{@{}lrrrrrr@{}}
\toprule
 & \multicolumn{3}{c}{Final submission} & \multicolumn{3}{c}{Verified new bests} \\
\cmidrule(lr){2-4}\cmidrule(l){5-7}
Policy & \shortstack[r]{Valid\\(\%)} & \shortstack[r]{Mean\\pct.} & \shortstack[r]{Pct.\ if\\valid} & \shortstack[r]{First\\(h)} & \shortstack[r]{Trials\\with one} & \shortstack[r]{Per\\trial} \\
\midrule
Base AIDE & 78.4 & 36.9 & 47.0 & 0.43 & 40 & 3.4 \\
\midrule
\multicolumn{7}{@{}l}{\emph{MLE-World guided}} \\
Sequential & 90.2 & 41.1 & 45.5 & 0.62 & 46 & 4.5 \\
\quad + confidence gate & 96.1 & 47.7 & 49.7 & 0.68 & 49 & 4.4 \\
Pairwise & 92.2 & 42.8 & 46.4 & 1.06 & 47 & 5.1 \\
\quad + confidence gate & 90.2 & 48.5 & 53.8 & 0.68 & 46 & 4.2 \\
Fusion & 96.1 & 47.2 & 49.1 & 0.56 & 49 & 16.5 \\
\quad + confidence gate & 94.1 & 46.4 & 49.3 & 0.75 & 48 & 7.1 \\
BoN & 88.2 & 39.5 & 44.8 & 0.93 & 45 & 6.9 \\
\quad + confidence gate & 94.1 & 47.2 & 50.2 & 0.58 & 48 & 6.3 \\
UCB & 88.2 & 45.1 & 51.1 & 0.57 & 45 & 11.3 \\
\quad + confidence gate & 90.2 & 49.2 & 54.6 & 0.39 & 46 & 5.9 \\
\bottomrule
\end{tabular}
\par\medskip
\begin{tabular}{@{}lrrrrrr@{}}
\toprule
 & \multicolumn{6}{c}{Real runs} \\
\cmidrule(l){2-7}
Policy & \shortstack[r]{Per\\trial} & \shortstack[r]{Clean\\(\%)} & \shortstack[r]{Crash\\(\%)} & \shortstack[r]{Time-\\out (\%)} & \shortstack[r]{Hours/\\trial} & \shortstack[r]{Timeout\\h/trial} \\
\midrule
Base AIDE & 144.9 & 13.4 & 83.8 & 2.8 & 6.65 & 3.33 \\
\midrule
\multicolumn{7}{@{}l}{\emph{MLE-World guided}} \\
Sequential & 47.8 & 34.1 & 60.0 & 5.9 & 5.23 & 2.25 \\
\quad + confidence gate & 21.6 & 52.6 & 39.6 & 7.8 & 3.62 & 1.33 \\
Pairwise & 63.9 & 28.5 & 67.3 & 4.2 & 5.41 & 2.04 \\
\quad + confidence gate & 17.1 & 56.8 & 35.3 & 7.9 & 2.91 & 0.98 \\
Fusion & 57.5 & 64.5 & 32.0 & 3.5 & 5.20 & 1.40 \\
\quad + confidence gate & 24.2 & 65.7 & 28.6 & 5.7 & 3.20 & 1.11 \\
BoN & 67.2 & 29.7 & 66.1 & 4.3 & 6.10 & 2.11 \\
\quad + confidence gate & 23.7 & 59.3 & 32.0 & 8.7 & 4.49 & 1.71 \\
UCB & 69.8 & 43.8 & 52.1 & 4.1 & 5.87 & 2.20 \\
\quad + confidence gate & 36.0 & 42.2 & 51.3 & 6.5 & 4.45 & 1.85 \\
\bottomrule
\end{tabular}

\end{table}

\paragraph{Why UCB produces fewer valid submissions.}
The valid only means of 51.1 for UCB and 49.1 for Fusion use different sets of trials.
On the 44 task--seed pairs valid under both policies, their means are 52.09 and 52.22.
Fusion's overall lead of 2.10 points decomposes into 0.12 points from these shared valid
pairs and 1.98 points from pairs where only one policy submits a valid result; these
contributions sum score differences over the respective pairs and divide by all 51 trials.

All six invalid ungated UCB trials end without a published submission: three on denoising dirty
documents, two on MLSP birds (seeds 34 and 36), and one on Jigsaw toxic comments (seed 35).
Their 978 committed steps comprise 497 Draft and 481 Debug steps, with none in Improve.
In 938 steps (95.9\%), no candidate has both predicted usability and an eligible metric;
selection therefore falls back to $p_{\mathrm{clean}}$. This score predicts exception free
execution, whereas publication also requires a valid CSV and an accepted validation metric.

The denoising trials repeatedly encounter image shape errors and malformed submissions
(missing identifiers, wrong row counts, or mismatched identifiers). Each produces one
format valid candidate that is rejected by metric review. The two bird trials suffer repeated
data parsing errors, including 178 recorded \texttt{ParserError}s. In the Jigsaw trial, six
timeouts consume 5.53 hours; its only exception free execution lacks a finite accepted metric
and receives an HTTP 500 response from the submission validator. Thus the recorded failures
include both program defects and a validation service error; reviewer rejection reasons are
recorded assessments, not independently verified diagnoses.

Fusion publishes a verified submission on five of these six corresponding task--seed pairs;
each first publication is a synthesized program. Fusion also fails on one pair where UCB
succeeds (plant pathology, seed 35), giving its net advantage of four valid submissions.
These records locate UCB's coverage deficit in obtaining and verifying an initial program.
They do not isolate a causal effect of the uncertainty bonus: it changes the pick relative to
metric greedy selection in only five of the 978 recorded decisions, and the policies follow
different trajectories.

\FloatBarrier
\section{Diagnostics on fixed states}
\label{app:fixed-candidates}

Table~\ref{tab:fixed-candidates} fixes the starting task state to isolate what MLE-World's predictions
contribute to a local decision, separately from the rest of the search. We sample 153 states from recorded agent runs,
three per task and stage (draft, debug, improve) on each of the 17 tasks, without consulting their
outcomes. At each state, Qwen3.5-9B generates four candidates and MLE-World predicts their
terminal outputs and structured fields. Selection, Pairwise, and Fusion use these same four candidates; Sequential
shares the first candidate and generates three revisions under each feedback condition. We execute
and grade every candidate and every program a rule produces; each rule's choice is recorded before
any private test grade is read. We omit UCB because these local outcomes do not measure the value
of exploration through later search decisions.

\paragraph{Model calls.}
The common pool contains four input programs. Pairwise adds three judge calls; Fusion adds one
generation call to synthesize a fifth program. Sequential instead adds three revision calls to the
shared first candidate and predicts all four attempts for final selection; it may also call a judge
when all four are predicted unusable. These are the nominal calls before generation retries.
Including code-parsing retries, each Fusion condition in Table~\ref{tab:fixed-candidates-compact}
makes 153 synthesis calls over 150 states (1.02 per state). Pairwise makes exactly three judge calls per
state. Predicted-feedback Pairwise and Fusion use four world-model predictions. Thus these comparisons
share task states and inputs, but do not match total model calls or execution costs.

\begin{table}[!t]
\caption{\textbf{Diagnostics on fixed states.} Screening and Fusion improve validity and test percentile;
the test percentile gains of Pairwise and Sequential from predicted terminal output are not established.
Selection, Pairwise, and Fusion share their input pool; Sequential shares
its starting candidate. \emph{Uniform}: expected
outcome of a random pick. \emph{Screen}: random pick among candidates MLE-World predicts usable with
a finite metric, else the largest $p_{\mathrm{clean}}(x_s)$. \emph{+ ranking}: the screened candidate
with the best predicted metric. \emph{Real validation}: best real validation score, else the
ranking's pick. Sequential uses the same selector by predicted metric and fallback in every condition;
\emph{Structured predictions} omit terminal output from revision feedback. Test pct.\ is zero if invalid. $\Delta$: paired
difference from the row above within each panel $\pm$ SE over 4,000 hierarchical task/run bootstrap
resamples. $^\dagger$Executes all four inputs. $^\ddagger$Uses real execution feedback during revision.}
\label{tab:fixed-candidates}
\centering
\begin{tabular}{@{}lrrrr@{}}
\toprule
Rule & \shortstack[r]{Mean\\percentile} & \shortstack[r]{$\Delta$ mean pct.\\(points)} & \shortstack[r]{Valid\\(\%)} & \shortstack[r]{$\Delta$ valid\\(points)} \\
\midrule
\multicolumn{5}{@{}l}{\emph{Selection: pick one of the same four candidates}} \\
Uniform over the pool & 10.9 & -- & 26.1 & -- \\
Screen by predicted outcome & 16.4 & +5.51\,{$\pm$1.18} & 41.9 & +15.8\,{$\pm$3.1} \\
\quad + ranking by predicted metric & 17.1 & +0.68\,{$\pm$0.79} & 42.5 & +0.5\,{$\pm$1.3} \\
Real validation of all four$^\dagger$ & 19.8 & +2.65\,{$\pm$2.19} & 47.1 & +4.6\,{$\pm$2.6} \\
\midrule
\multicolumn{5}{@{}l}{\emph{Pairwise: judge the same four candidates}} \\
Code only & 13.0 & -- & 32.7 & -- \\
Predicted terminal output & 15.6 & +2.60\,{$\pm$1.89} & 37.9 & +5.2\,{$\pm$3.6} \\
Real terminal output$^\dagger$ & 14.6 & $-$1.03\,{$\pm$1.46} & 34.6 & $-$3.3\,{$\pm$2.2} \\
\midrule
\multicolumn{5}{@{}l}{\emph{Sequential: refine the same starting candidate}} \\
No revision feedback & 10.4 & -- & 28.0 & -- \\
Structured predictions & 12.3 & +1.86\,{$\pm$2.16} & 31.2 & +3.2\,{$\pm$4.5} \\
Predicted terminal output & 15.3 & +2.99\,{$\pm$2.50} & 38.0 & +6.8\,{$\pm$4.3} \\
Real terminal output$^\ddagger$ & 15.0 & $-$0.26\,{$\pm$3.48} & 39.4 & +1.4\,{$\pm$4.9} \\
\midrule
\multicolumn{5}{@{}l}{\emph{Fusion: synthesize from the same four candidates}} \\
Code only & 14.0 & -- & 36.7 & -- \\
Predicted terminal output & 20.6 & +6.55\,{$\pm$2.87} & 49.2 & +12.5\,{$\pm$4.5} \\
Real terminal output$^\dagger$ & 20.4 & $-$0.21\,{$\pm$1.71} & 46.5 & $-$2.7\,{$\pm$4.3} \\
\bottomrule
\end{tabular}

\end{table}

\paragraph{Screening and Fusion improve immediate outcomes.}
To isolate the contribution of predictions, we fix the task state and its four generated candidates,
execute and grade every candidate, and vary only how the predictions are used
(Table~\ref{tab:fixed-candidates}). Screening and Fusion show the clearest gains in this controlled
comparison. First, the predictions prevent failed runs: choosing
among the candidates MLE-World predicts will run raises the share of chosen programs that yield a
valid submission from 26.1\% to 41.9\% ($+15.8$ points), and predicted feedback raises
fusion's share from 36.7\% to 49.2\% ($+12.5$ points). As in complete runs, the chosen
program executes about as long (256 versus 260 seconds), but more of that time goes to runs that
finish cleanly (135 versus 96 seconds; $+39$ seconds).
Second, test percentile rises by 5.51 and 6.55 points, mostly through this higher validity.
Subsequent metric ranking adds only 0.68 points after screening.
Fusion's observed means with predicted and real feedback are 20.6 and 20.4 percentile.

\paragraph{Selection.}
Each rule picks one of the same four candidates. The screen picks uniformly among candidates
predicted usable with a finite predicted metric and, when none qualifies, the candidate with the
largest $p_{\mathrm{clean}}(x_s)$. No candidate qualifies in 80 of the 153 states. In the 73 states where the
screen keeps at least one candidate, it gains 9.16 percentile points; the
fallback states gain 2.29 points. Ranking the screened candidates by predicted metric
adds little. Selecting by the real validation scores of all four, which requires executing every
candidate, reaches 19.8, 2.65 points above ranking by predicted metric.

\paragraph{Pairwise selection.}
The same four candidates are compared in pairs by a judge that receives code only,
predicted terminal output, or real terminal output. Predicted terminal output raises test percentile by 2.60 points
over code only, which is not established; validity rises from 32.7\% to 37.9\%.

\paragraph{Sequential refinement.}
Each condition starts with the same first candidate and generates three revisions, using no
revision feedback, structured predictions without terminal output, predicted terminal output, or real
terminal output. All conditions select among their four attempts by predicted metric with the same fallback
rule, so the intervention changes revision feedback rather than final selection. Predicted
terminal output raises test percentile by 4.85 points over no feedback and by 2.99 over
structured predictions; neither gain is established.

\paragraph{Fusion.}
Fusion combines the same four candidates into a new program given code only,
MLE-World's predicted feedback, or real feedback from executing all four. Predicted feedback raises
the rate of valid submissions by 12.5 points and test percentile by 6.55, and real terminal output scores no
higher ($-0.21$ points). Three states are excluded from every Sequential and Fusion
condition because their runs did not complete (one exceeded the generator's context limit and two
were stopped during execution); the exclusion uses completion status alone. Letting the three
missing paired differences take any value from $-100$ to $+100$ bounds Fusion's gain from predicted feedback
between $+4.52$ and $+8.44$ points.

\paragraph{Real terminal output does not establish a further local gain.}
Replacing predicted with real terminal output changes test percentile by $-1.03$ points for Pairwise
(paired 95\% interval $[-4.05,1.80]$), $-0.26$ for Sequential ($[-7.28,6.44]$), and $-0.21$
for Fusion ($[-3.45,3.36]$). Selecting by real validation scores instead of predicted metrics
gains $+2.65$ points, also uncertain ($[-0.24,8.01]$). These comparisons find no established
gain from substituting real feedback in the tested local decisions. They do not establish
equivalence between predicted and real feedback or rule out benefits over longer searches.

\paragraph{Fusion leads the fixed state comparison.}
On the 150 states shared by all four local methods, Fusion has the highest mean percentile (20.6),
followed by greedy selection by metric (17.3), Pairwise (15.4), and Sequential (15.3). Fusion exceeds
Pairwise by 5.22 points (paired 95\% interval $[1.56,10.04]$) and Sequential by 5.31
($[0.26,11.77]$). Pairwise and Sequential differ by just 0.09 points ($[-4.31,4.62]$).
Fusion also exceeds greedy selection by metric in mean quality by 3.28 points, although this
interval includes zero ($[-0.09,8.30]$). These comparisons support Fusion's advantage over
Pairwise and Sequential; the higher mean relative to metric selection is descriptive.

\paragraph{Relation to production BoN.}
The 19.8 percentile real validation reference in Table~\ref{tab:fixed-candidates} additionally
executes all four inputs, whereas production BoN executes only its selected program.
The diagnostic's predicted metric rule is also not the full BoN policy: when no candidate qualifies,
it uses the shared $p_{\mathrm{clean}}(x_s)$ fallback. This occurs in 80 of the 153 pools.
Production BoN instead returns the first usable candidate or invokes the failed pool judge
(Algorithm~\ref{alg:tts-bon}). Moreover, the diagnostic measures one local output at a fixed
starting state, with different costs of model calls, rather than the final submission after
eight hours of changing candidate pools and search states.

\paragraph{Where the eight hour BoN--Fusion gap appears.}
Fusion reaches 47.2 percentile versus BoN's 39.5 in
Table~\ref{tab:ungated-agent-performance}. Both final validity (96.1\% versus 88.2\%) and mean
percentile among valid submissions (49.1 versus 44.8) are higher; these conditional means use
each policy's own valid trials (Table~\ref{tab:agent-execution}).
Fusion also records more verified new bests per trial (16.5 versus 6.9) despite fewer committed
search steps (59.8 versus 68.3), and a larger share of its real runs finish cleanly (64.5\% versus
29.7\%). These observations are consistent with more productive candidate generation and
refinement. They do not identify which mechanism causes the final quality gap: attributing it
to repeated synthesis or fallback behavior requires a matched intervention over complete searches.

\section{Choice of primary agent metric}
\label{app:medal-metric}

MLE-bench compares each final submission with the competition's human leaderboard. Its medals use
Kaggle's thresholds: gold requires the top 10\% of teams on leaderboards with fewer than 100 teams,
the top 10 on leaderboards with 100\rng{}249 teams, and the top 10 plus 0.2\% of all teams
(rounded down) otherwise. Ten of our 17
leaderboards have more than 1,000 teams, where gold means roughly the top 1\%. Leaderboard
percentile instead measures the share of entries a submission beats.
We use percentile as the primary metric because medal rates are all or nothing: 12 of our 17 tasks
never yield a gold, while one task with a small leaderboard yields more than half of all golds. \citet{aira2}
likewise rank agents by percentile because medal thresholds amplify noise and hide progress on
hard tasks, and \citet{rpm} report a continuous normalized score.
Tables~\ref{tab:medal-tasks} and~\ref{tab:medal-policies} break down the 306 final submissions
of the six earlier 8\,h ungated policies by task and by policy.

\paragraph{One task supplies more than half of all golds.}
Detecting-insults has a leaderboard of 50 teams, so gold means the top five. Seventeen of its 18
submissions win gold; only one Fusion seed misses. The task accounts for 17 of the 31 golds and
adds up to three golds, or 5.9 points of gold rate, to every policy without separating them.

\paragraph{Most tasks never produce a medal.}
Twelve tasks yield no gold in any submission and nine yield no medal of any kind, so medal rates
ignore them. Percentile still separates policies on these tasks. Means for each policy range from 0.0
to 36.6 on mlsp-2013-birds, where three policies never produce a valid submission, and from 2.9
to 53.9 on English text normalization, where a single submission earns a medal.

\paragraph{Where golds occur, a single seed decides them.}
On the other four gold tasks, gold requires the 98.7th to 99.1st percentile or the best
leaderboard score. On aerial-cactus, 150 teams share a perfect AUC of 1.0, so gold requires
exactly 1.0: a submission scoring 0.99996 ranks at the 87.7th percentile and wins no medal. With
three seeds per task, one seed moves a policy's gold rate by 2.0 points.

\paragraph{Gold rate can reverse the ranking.}
Base AIDE has twice Fusion's gold rate (11.8\% versus 5.9\%; paired difference $-5.9$ points),
yet Fusion submits valid solutions more often (96.1\% versus
78.4\%) and reaches a higher mean percentile ($+10.4$ points). Pairwise exceeds Fusion in
gold rate ($+9.8$ points) but trails it in mean percentile ($-4.4$ points).
Table~\ref{tab:medal-policies} shows why. Fusion's three golds contribute 5.9 of its 47.2
percentile points, and its 46 other valid submissions, at a mean 45.8th percentile, contribute
41.3. Pairwise's eight golds contribute 15.4 of its 42.8 points, but its 39 other valid
submissions average only the 35.8th percentile, and its four invalid submissions score zero.

\paragraph{Percentile has its own limit.}
Where leaderboard scores cluster, small score gains move percentile sharply: on English text
normalization, accuracy 0.964 ranks at the 6th percentile and 0.990 at the 47th. Averaging over 17
equally weighted tasks limits the influence of any one leaderboard. We therefore base conclusions on mean percentile and
report medal rates only for comparison with MLE-bench.

\begin{table}[!htbp]
\caption{\textbf{Medals by task.} More than half of all golds come from one task and twelve tasks yield
none, yet policies still differ in percentile on tasks without medals. The 306 final submissions
of the six earlier 8\,h ungated policies (18 per task: six
policies, three seeds). Teams:
leaderboard size. Gold rank: the rank a submission must reach for gold. Gold pct.: percentile of a
submission scoring exactly the gold threshold. Golds and Any medal count submissions. Policy pct.:
lowest and highest of the six mean percentiles of the policies. $^\dagger$Gold requires the best
leaderboard score, which many teams share: AUC 1.0 on aerial-cactus (150 teams) and log loss 0.0
on leaf-classification (24 teams).}
\label{tab:medal-tasks}
\centering
\begin{tabularx}{\linewidth}{@{}>{\raggedright\arraybackslash}Xrrrrrr@{}}
\toprule
Task & Teams & \shortstack[r]{Gold\\rank} & \shortstack[r]{Gold\\pct.} & Golds & \shortstack[r]{Any\\medal} & \shortstack[r]{Policy pct.\\(range)} \\
\midrule
\multicolumn{7}{@{}l}{\emph{Tasks with a gold}} \\
\texttt{detecting-\allowbreak insults-\allowbreak in-\allowbreak social-\allowbreak commentary} & 50 & 5 & 90.0 & 17 & 17 & 81.3\rng{}100.0 \\
\texttt{dogs-\allowbreak vs-\allowbreak cats-\allowbreak redux-\allowbreak kernels-\allowbreak edition} & 1,315 & 12 & 99.1 & 4 & 9 & 33.3\rng{}95.5 \\
\texttt{plant-\allowbreak pathology-\allowbreak 2020-\allowbreak fgvc7} & 1,318 & 12 & 99.1 & 4 & 5 & 10.1\rng{}92.9 \\
\texttt{tabular-\allowbreak playground-\allowbreak series-\allowbreak dec-\allowbreak 2021} & 1,189 & 12 & 98.7 & 4 & 4 & 46.3\rng{}72.7 \\
\texttt{aerial-\allowbreak cactus-\allowbreak identification} & 1,221 & 12 & 87.7$^\dagger$ & 2 & 2 & 50.4\rng{}69.8 \\
\midrule
\multicolumn{7}{@{}l}{\emph{Medals, but no gold}} \\
\texttt{nomad2018-\allowbreak predict-\allowbreak transparent-\allowbreak conductors} & 879 & 11 & 98.7 & 0 & 15 & 62.7\rng{}96.1 \\
\texttt{random-\allowbreak acts-\allowbreak of-\allowbreak pizza} & 462 & 10 & 97.8 & 0 & 14 & 68.2\rng{}87.7 \\
\texttt{text-\allowbreak normalization-\allowbreak challenge-\allowbreak english-\allowbreak language} & 261 & 10 & 96.2 & 0 & 1 & 2.9\rng{}53.9 \\
\midrule
\multicolumn{7}{@{}l}{\emph{No medal}} \\
\texttt{spooky-\allowbreak author-\allowbreak identification} & 1,242 & 12 & 99.0 & 0 & 0 & 40.1\rng{}55.4 \\
\texttt{leaf-\allowbreak classification} & 1,596 & 13 & 98.5$^\dagger$ & 0 & 0 & 17.8\rng{}47.9 \\
\texttt{dog-\allowbreak breed-\allowbreak identification} & 1,281 & 12 & 99.1 & 0 & 0 & 17.6\rng{}42.8 \\
\texttt{tabular-\allowbreak playground-\allowbreak series-\allowbreak may-\allowbreak 2022} & 1,152 & 12 & 98.4 & 0 & 0 & 27.5\rng{}36.6 \\
\texttt{mlsp-\allowbreak 2013-\allowbreak birds} & 81 & 8 & 90.1 & 0 & 0 & 0.0\rng{}36.6 \\
\texttt{new-\allowbreak york-\allowbreak city-\allowbreak taxi-\allowbreak fare-\allowbreak prediction} & 1,485 & 12 & 99.2 & 0 & 0 & 5.3\rng{}25.3 \\
\texttt{denoising-\allowbreak dirty-\allowbreak documents} & 162 & 10 & 93.8 & 0 & 0 & 0.0\rng{}22.2 \\
\texttt{jigsaw-\allowbreak toxic-\allowbreak comment-\allowbreak classification-\allowbreak challenge} & 4,539 & 19 & 99.6 & 0 & 0 & 7.0\rng{}20.8 \\
\texttt{text-\allowbreak normalization-\allowbreak challenge-\allowbreak russian-\allowbreak language} & 163 & 10 & 93.9 & 0 & 0 & 9.8\rng{}19.2 \\
\midrule
All 17 tasks & & & & 31 & 67 & \\
\bottomrule
\end{tabularx}

\end{table}

\begin{table}[!htbp]
\caption{\textbf{Medals and percentile by policy.} Gold counts turn on a few submissions, whereas
most of each policy's mean percentile comes from submissions without a gold. Counts over 51
trials per policy. Mean percentile (Total, as in Table~\ref{tab:ungated-agent-performance})
splits into points from gold medal trials and from other valid trials; invalid submissions
contribute zero, and rounded components may not sum to the total.}
\label{tab:medal-policies}
\centering
\begin{tabular}{@{}lrrrrrrr@{}}
\toprule
 & \multicolumn{2}{c}{Golds} & & & \multicolumn{3}{c}{Mean percentile} \\
\cmidrule(lr){2-3}\cmidrule(l){6-8}
Policy & \shortstack[r]{detecting-\\insults} & Other & \shortstack[r]{Any\\medal} & Invalid & \shortstack[r]{Gold\\trials} & \shortstack[r]{Other valid\\trials} & Total \\
\midrule
Base AIDE & 3 & 3 & 11 & 11 & 11.8 & 25.1 & 36.9 \\
Sequential & 3 & 3 & 11 & 5 & 11.5 & 29.5 & 41.1 \\
Pairwise & 3 & 5 & 14 & 4 & 15.4 & 27.4 & 42.8 \\
Fusion & 2 & 1 & 11 & 2 & 5.9 & 41.3 & 47.2 \\
BoN & 3 & 0 & 9 & 6 & 5.9 & 33.6 & 39.5 \\
UCB & 3 & 2 & 11 & 6 & 9.8 & 35.3 & 45.1 \\
\bottomrule
\end{tabular}

\end{table}

\FloatBarrier
\section{Execution prediction on tasks excluded from finetuning}
\label{app:ood}

\paragraph{Setup.}
We evaluate execution prediction on APTOS Blindness Detection, 3D Object Detection,
and RANZCR Catheter/Line Classification, all excluded from finetuning.
All methods use the same 300 programs: 100 per task, sampled uniformly without
replacement within stage from each task's 200-program confirmation set
(34 Draft, 33 Improve, 33 Debug).
An audit of all 497,056 finetuning examples found neither these task identities nor
exact program matches after whitespace normalization.

\begin{table}[!htbp]
\centering
\caption{\textbf{Exploratory prediction on three tasks excluded from finetuning.}
Accuracy (\%, $\pm$ program bootstrap SE) on 300 programs, 100 per task with balanced
agent stages. The fields and scoring match Table~\ref{tab:wm-fidelity}; DeepSeek uses
default reasoning effort here. Bold denotes the best value in each column.}
\label{tab:ood-fidelity}
\begingroup

\begin{tabular}{@{}lrrrr@{}}
\toprule
Method & \shortstack[r]{Terminal\\output} & \shortstack[r]{Exception\\class} & \shortstack[r]{Crash\\class} & \shortstack[r]{Validation\\metric} \\
\midrule
DeepSeek & 42.1{$\pm$2.8} & \textbf{63.7}{$\pm$2.8} & 45.4{$\pm$3.2} & 21.3{$\pm$5.8} \\
Coder & 21.1{$\pm$2.2} & 30.3{$\pm$2.6} & 18.9{$\pm$2.4} & 31.9{$\pm$6.8} \\
\textbf{MLE-World (4B, no CoT)} & \textbf{53.3}{$\pm$2.8} & 60.3{$\pm$2.8} & \textbf{56.3}{$\pm$3.1} & \textbf{44.7}{$\pm$7.3} \\
\bottomrule
\end{tabular}
\endgroup

\end{table}

\paragraph{Potential transfer mechanisms.}
New tasks can reuse familiar execution operations: loading images, transforming
tensors, and adapting pretrained models. Failures in these shared operations may
remain predictable even when the data and prediction target change
(Table~\ref{tab:ood-relations}). Task-specific geometry, labels, and metrics introduce
less familiar behavior. Shared code therefore offers a plausible route to transfer,
though training family ablations are needed to establish its role.

\begin{table}[!htbp]
\centering
\caption{\textbf{Shared operations that could support transfer.} The middle column names
tasks present in finetuning. These are proposed mechanisms, not measured causal effects.}
\label{tab:ood-relations}

\begin{tabular}{@{}>{\raggedright\arraybackslash}p{0.21\linewidth}>{\raggedright\arraybackslash}p{0.26\linewidth}>{\raggedright\arraybackslash}p{0.47\linewidth}@{}}
\toprule
Held out task & Related training tasks & Shared structure and remaining shift \\
\midrule
APTOS Blindness Detection & Plant Pathology 2020; Dog Breed & Image classification with
pretrained visual encoders; retinal severity introduces ordinal labels and a new metric. \\
3D Object Detection & Dog Breed; Aerial Cactus & Shared image, array, and PyTorch
components can expose familiar import and file errors; 3D geometry and detection remain
new task specific computations. \\
RANZCR Catheter/Line & Plant Pathology 2020 & Image preprocessing and visual prediction
heads are shared; chest radiographs and multiple catheter/line labels change the target. \\
\bottomrule
\end{tabular}
\end{table}

\paragraph{Results and limitations.}
The pooled and per-task results (Tables~\ref{tab:ood-fidelity}
and~\ref{tab:ood-fidelity-by-task}) suggest that execution prediction can transfer
to tasks excluded from finetuning, with gains that vary across output fields.
Because failed executions dominate the sample, the evidence mainly concerns
recognizing execution failures on new tasks. Predicting successful training outcomes
and timeouts remains less established. A broader, prospectively selected evaluation
and training family ablations would help distinguish transfer through shared code
from generalization across task families and software environments.

\begin{table}[!htbp]
\centering
\caption{\textbf{Per task prediction accuracy on the three selected tasks.}
Accuracy (\%, $\pm$ program bootstrap SE) using the same methods and metrics as
Table~\ref{tab:ood-fidelity}. Each task has 100 programs: 34 Draft, 33 Improve, and 33 Debug.
Gray $\dagger$ cells have fewer than ten metric references;
SE is omitted when the metric is undefined in any bootstrap resample.}
\label{tab:ood-fidelity-by-task}
\begingroup

\begin{tabular}{@{}lrrrr@{}}
\toprule
Method & \shortstack[r]{Terminal\\output} & \shortstack[r]{Exception\\class} & \shortstack[r]{Crash\\class} & \shortstack[r]{Validation\\metric} \\
\midrule
\multicolumn{5}{@{}l}{\emph{APTOS Blindness Detection}} \\
DeepSeek & 49.3{$\pm$4.9} & \textbf{68.0}{$\pm$4.5} & \textbf{59.0}{$\pm$5.6} & 16.7{$\pm$8.6} \\
Coder & 18.3{$\pm$3.8} & 32.0{$\pm$4.6} & 15.3{$\pm$4.2} & 16.7{$\pm$9.1} \\
\textbf{MLE-World (4B, no CoT)} & \textbf{53.0}{$\pm$4.9} & 60.0{$\pm$4.8} & 53.6{$\pm$5.6} & \textbf{50.0}{$\pm$12.1} \\
\addlinespace[3pt]
\multicolumn{5}{@{}l}{\emph{3D Object Detection}} \\
DeepSeek & 35.7{$\pm$4.7} & 52.0{$\pm$4.9} & 26.5{$\pm$5.0} & 31.8{$\pm$10.1} \\
Coder & 30.0{$\pm$4.6} & 30.0{$\pm$4.6} & 25.6{$\pm$5.1} & 45.5{$\pm$10.7} \\
\textbf{MLE-World (4B, no CoT)} & \textbf{48.7}{$\pm$5.1} & \textbf{53.0}{$\pm$5.1} & \textbf{47.0}{$\pm$5.8} & \textbf{50.0}{$\pm$10.9} \\
\addlinespace[3pt]
\multicolumn{5}{@{}l}{\emph{RANZCR Catheter/Line}} \\
DeepSeek & 41.3{$\pm$4.9} & \textbf{71.0}{$\pm$4.4} & 51.0{$\pm$5.4} & {\color{gray}0.0$^{\dagger}$} \\
Coder & 15.0{$\pm$3.5} & 29.0{$\pm$4.6} & 15.7{$\pm$4.0} & {\color{gray}\textbf{28.6}$^{\dagger}$} \\
\textbf{MLE-World (4B, no CoT)} & \textbf{58.3}{$\pm$4.7} & 68.0{$\pm$4.6} & \textbf{67.5}{$\pm$5.0} & {\color{gray}14.3$^{\dagger}$} \\
\addlinespace[3pt]
\bottomrule
\end{tabular}
\endgroup

\end{table}

\end{document}